%% file: PP-BD-WWSN.tex
\documentclass[review]{elsarticle}
\usepackage{indentfirst}
\usepackage{latexsym}
\usepackage{lipsum}
\usepackage{filecontents}
\usepackage[english]{babel}
\usepackage{epstopdf}
\usepackage{graphicx}
\graphicspath{{figures/}}
\usepackage{tabularx}
\usepackage{threeparttable}
\usepackage{booktabs}
\usepackage{enumerate}
\usepackage{paralist}
\usepackage[svgnames]{xcolor}
\usepackage{graphicx} 			
\usepackage{subcaption}	
\usepackage{verbatim}
\usepackage{xspace,xcolor,amsmath,mathtools,url,amsfonts,amssymb}
\usepackage{wrapfig,framed,esvect,microtype,tikz}
\usepackage{protocolj}
\usepackage{dsfont}
\usepackage{amsmath,scalerel}
\usepackage{amsthm} 
\usepackage{textcomp} 
\usepackage{savesym}
\usepackage{amsmath}
\usepackage{varwidth} 
\usepackage{float}
\usepackage{subfig}               
\usepackage{overpic}

\usepackage{mathrsfs}

\usepackage{multicol}
\usepackage{color}
\usepackage[ruled,linesnumbered,procnumbered]{algorithm2e}
\usepackage{makecell}
\usepackage{lineno}
\usepackage{soul}
\usepackage{multirow}
\usepackage{geometry}
\usepackage{ulem}

\usepackage{epsfig}
\usepackage{threeparttable}
\usepackage{textcomp}
\usepackage{fancybox}
\usepackage{amssymb}
\usepackage{mathtools}
\usepackage{pifont}
\usepackage{xcolor}
\input{header/header-Fixed}

\input{header/header-diy}
\input{header/header-MAG}
\journal{Computers \& Security}

\begin{document}

\begin{frontmatter}

\title{GhostEncoder: Stealthy Backdoor Attacks with Dynamic Triggers to Pre-trained Encoders in Self-supervised Learning}

\author[1]{Qiannan Wang}
\ead{qnwang@nuaa.edu.cn}
\author[1,2]{Changchun Yin}
\ead{ycc0801@nuaa.edu.cn}
\author[3]{Zhe Liu}
\ead{zhe.liu@zhejianglab.edu.cn}
\author[1,2]{Liming Fang}
\ead{fangliming@nuaa.edu.cn}
\author[4]{Run Wang}
\ead{wangrun@whu.edu.cn}
\author[5]{Chenhao Lin}
\ead{linchenhao@xjtu.edu.cn}

\address[1]{College of Computer Science and Technology, Nanjing University of Aeronautics and Astronautics, Nanjing, China}
\address[2]{Nanjing University of
Aeronautics and Astronautics
Shenzhen Research Institute, Shenzhen, China}
\address[3]{Research Institute of Basic Theories, Zhejiang Lab, Hangzhou, China}
\address[4]{School of Cyber Science and Engineering, Wuhan University, Wuhan, China}
\address[5]{Faculty of
Electronic and Information Engineering, Xi'an Jiaotong University, Xi'an, China}
\cortext[cor1]{Changchun Yin is the Corresponding author.}

\begin{abstract}
 Within the realm of computer vision, self-supervised learning (SSL) pertains to training pre-trained image encoders utilizing a substantial quantity of unlabeled images. Pre-trained image encoders can serve as feature extractors, facilitating the construction of downstream classifiers for various tasks. However, the use of SSL has led to an increase in security research related to various backdoor attacks. Currently, the trigger patterns used in backdoor attacks on SSL are mostly visible or static (sample-agnostic), making backdoors less covert and significantly affecting the attack performance. In this work, we propose GhostEncoder, the first dynamic invisible backdoor attack on SSL. Unlike existing backdoor attacks on SSL, which use visible or static trigger patterns, GhostEncoder utilizes image steganography techniques to encode hidden information into benign images and generate backdoor samples. We then fine-tune the pre-trained image encoder on a manipulation dataset to inject the backdoor, enabling downstream classifiers built upon the backdoored encoder to inherit the backdoor behavior for target downstream tasks. We evaluate GhostEncoder on three downstream tasks and results demonstrate that GhostEncoder provides practical stealthiness on images and deceives the victim model with a high attack success rate without compromising its utility. Furthermore, GhostEncoder withstands state-of-the-art defenses, including STRIP, STRIP-Cl, and SSL-Cleanse. 
\end{abstract}


 \end{frontmatter}

\section{Introduction}
Self-supervised learning (SSL) is an emerging paradigm in machine learning that not only reduces reliance on labeled data but also enhances adversarial robustness by introducing challenges to manipulating model predictions\cite{grill2020bootstrap,he2020momentum,hjelm2018learning,9,chen2020improved}. However, SSL's performance heavily depends on abundant unlabeled data, which increases the associated computational costs. Regular users tend to adopt third-party pre-trained encoders available online to mitigate this issue. Yet, the opacity of the training process introduces new security threats. Backdoor attacks \cite{gu2017badnets} represent a menace within Deep Neural Network (DNNs) training, involving the malicious manipulation of prediction behavior through the contamination of a subset of training samples. Attackers implant hidden backdoors by embedding selected patterns (referred to as backdoor triggers) into images, creating a backdoor training set, and training the model with this set \cite{liu2018trojaning,yao2019latent,carlini2021poisoning}. The backdoored model will covertly alter predictions to target label when the trigger is embedded in inputs while maintaining normal behavior on benign samples.

\begin{figure}[t]
    \centering
    \includegraphics[width=0.6\columnwidth]{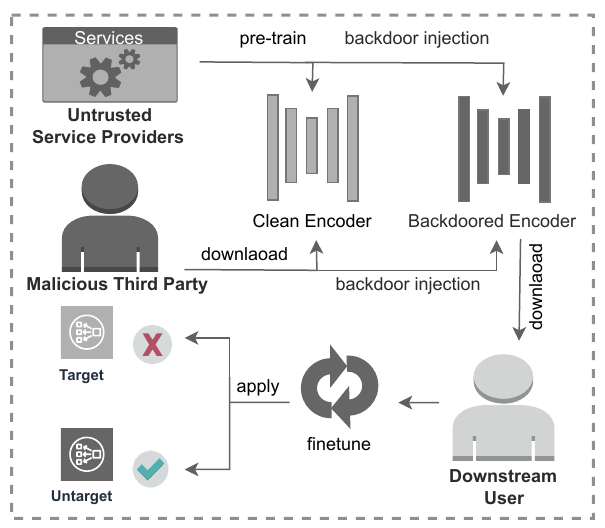}
    \caption{The overview framework of the proposed method.}
    \label{fig:service}
\end{figure}

Pre-trained encoders are employed as feature extractors for downstream classifiers in SSL. However, obtaining completely trustworthy encoders is an exceedingly challenging issue. As depicted in Figure \ref{fig:service}, one form of attacker involves an untrusted service provider who injects backdoors into their pre-trained encoders, which are subsequently provided to users for fine-tuning in downstream tasks. Consequently, the backdoor becomes active in the attacker's targeted downstream tasks, while the performance on non-targeted downstream tasks remains unaffected. Malicious third parties constitute another type of attacker. They download untainted image encoders, inject backdoors into them, and then release the compromised encoders on open-source platforms such as GitHub, Hugging Face, and Model Zoo, which makes these encoders available for unsuspecting victims to download. In particular, Liu et al.~\cite{1} introduced a method that involves modifying model weights to inject backdoors into pre-trained encoders, associating trigger patterns with reference inputs from the representation space of the target class. Furthermore, there exist poison-based attack methods, assuming adversaries can contaminate a small portion of training data without exerting any control over the training process. Saha et al.~\cite{saha2022backdoor} proposed poison-based backdoor attacks on SSL. Novel approaches have also adopted this poison-based attack strategy~\cite{li2022demystifying}, albeit with alterations limited to the trigger.

However, the concealment effectiveness of the trigger patterns used in these existing backdoor attacks is limited, as their backdoor triggers are sample-agnostic. This means that different backdoored samples contain the same trigger, regardless of which trigger pattern is used. As the trigger is sample-agnostic, defenders can easily reconstruct or detect the backdoor triggers based on shared behaviors among different backdoored samples, making the backdoor easily mitigated or eliminated by current defenses. For instance, in supervised scenarios, STRIP\cite{strip} is effective in defending against patch-based backdoor attacks with fixed triggers. However, we observe its limited efficacy in self-supervised scenarios. Then we reveal the reasons behind the failure of STRIP in the self-supervised setting and propose a new defense called STRIP-Cl. The experiments demonstrate that STRIP-Cl can significantly mitigate patch-type trigger-based backdoor attacks.

In this work, we propose a novel dynamic stealthy backdoor attack on SSL, called GhostEncoder, which can evade all existing defenses. GhostEncoder utilizes dynamic backdoor triggers that are specific to individual samples, resulting in diverse triggers for different backdoored samples. Importantly, perturbations are crafted to be globally imperceptible to human perception. Inspired by DNN-based image steganography \cite{baluja2017hiding,tancik2020stegastamp,zhu2018hidden}, we employ a pre-trained encoder-decoder network to encode an attacker-specified string into benign images. This encoding process generates sample-specific invisible additive noise, which serves as the backdoor trigger. The trigger is then seamlessly embedded into clean images, giving rise to backdoored images. Subsequently, the clean encoder undergoes fine-tuning to incorporate backdoors. This is achieved by bringing backdoored images closer to a reference input from the targeted class of a downstream task within the feature space. This enables the model to establish a connection between the trigger and the targeted class, facilitating effective backdoor learning.

We conduct a comprehensive evaluation of GhostEncoder on two distinct pre-training datasets, spanning across three diverse downstream tasks. Experimental results demonstrate that GhostEncoder achieves a high attack success rate (ASR) while minimizing the impact on the accuracy of downstream classifiers. Notably, in a scenario where the clean image encoder is fine-tuned on CIFAR10 and the backdoor downstream classifier is constructed on STL10, GhostEncoder attains an impressive ASR of 96.57$\%$. This success is attributed to the dynamic and unique trigger pattern introduced by GhostEncoder, which defies the underlying assumptions of certain existing defense mechanisms that anticipate the usage of fixed trigger patterns by attackers. As a result, GhostEncoder effortlessly circumvents such defenses, including our own robust defense proposal, STRIP-Cl. Furthermore, we investigate the effectiveness of GhostEncoder against two contemporary state-of-the-art defense approaches, namely DECREE \cite{feng2023detecting} and SSL-cleanse \cite{zheng2023ssl}. While these defenses exhibit the capability to recover the trigger from backdoor images and offer a level of resilience against the attack, their success relies on the premise of a consistent trigger pattern employed by the attacker. In contrast, GhostEncoder's utilization of global, sample-specific perturbations renders these defenses ineffective in countering its dynamic attack strategy.

\subsection{Contribution}
Our key contributions are summarized as follows:
\begin{itemize}
\item We propose GhostEncoder, the first dynamic invisible backdoor attack on SSL, significantly enhancing the robustness of the attack.

\item We investigate the reasons for the failure of STRIP in self-supervised scenarios and propose a novel enhancement method called STRIP-CL. It improves the defense capabilities of the STRIP method and expands its defense scope, effectively countering backdoor attacks that use patch-type trigger patterns.

\item We systematically evaluate GhostEncoder on multiple upstream datasets and multiple downstream tasks to demonstrate the effectiveness of the attack.

\item We explore three defense mechanisms, namely SSL-cleanse and DECREE, to mitigate the impact of GhostEncoder. However, our experimental results emphasize the need for new defense approaches to effectively counter GhostEncoder.
\end{itemize}

\section{Preliminaries}

\subsection{Image Steganography}
Image steganography is a sophisticated method that involves concealing confidential data within an image, aiming to maintain its imperceptibility to observers while minimizing any discernible alterations to the original image. The concealable amount of information is typically quantified in bits per pixel (bpp), where a lower bpp value signifies a reduced level of concealed data to mitigate the risk of detection. However, despite these efforts, advanced statistical analyses and detection techniques can potentially uncover these covert messages. In response to this challenge, the research focus has shifted toward leveraging Deep Neural Networks (DNNs) as a solution. DNNs offer heightened embedding capacity and enhanced security measures to counteract these vulnerabilities.

In their work, the authors propose a new method for image steganography using DNNs \cite{baluja2017hiding}. Their methodology entails the training of two interconnected networks: a hiding network and a revealing network, which collectively facilitate the embedding and extraction of full-size color images within other images of equivalent dimensions. The hiding network operates by taking a cover image and a transformed secret image as input, subsequently generating a container image. To prepare the secret image for processing, a pre-network operation is performed, which involves resizing the image if needed and converting its color-based pixels into more suitable feature representations, such as edges. Conversely, the revealing network is responsible for decoding the container image, unveiling the concealed secret image.

During the training process, the authors minimize the following error terms between the cover and secret images for reconstruction:

\begin{equation}
 \mathcal{L}\left(\mathbf{c}, \mathbf{c}^{\prime}, \mathbf{s}, \mathbf{s}^{\prime}\right)=\left\|\mathbf{c}-\mathbf{c}^{\prime}\right\|+\beta\left\|\mathbf{s}-\mathbf{s}^{\prime}\right\|   
\end{equation}
where $\mathbf{c}$ and $\mathbf{s}$ represent the cover and secret images, $\mathbf{c}^{\prime}$ and $\mathbf{s}^{\prime}$ represent the reconstructed images, and $\beta$ is the weight that balances the importance of each error term. It is worth noting that the error term $\left\|\mathbf{c}-\mathbf{c}^{\prime}\right\|$ is not applicable to the weights of the Reveal Network, ensuring that the information of the container image does not affect the extraction of the secret image.

To prevent the network from relying solely on LSBs, a small amount of noise is added to the container image during the training process. This noise occasionally flips LSBs, preventing them from being the exclusive carriers for reconstructing the secret image. This method extends the principles of traditional steganography methods by leveraging the capabilities of DNNs to distribute the secret image's representation across multiple bits of the carrier image. By doing so, this approach transcends the constraints of conventional LSBs operations and achieves a greater embedding capacity.

\subsection{Self-supervised Learning}

SSL has become a popular approach in the field of machine learning and computer vision because it allows learning representations from unlabeled data. By learning to predict certain attributes or relationships in the data, the model becomes capable of capturing valuable and meaningful representations that can subsequently be leveraged for various downstream tasks. The SSL pipeline typically consists of two stages: pre-training an image encoder and constructing a downstream classifier, which is then fine-tuned.

Among many methods, contrastive learning has demonstrated remarkable performance, such as MoCo \cite{8}, SimCLR \cite{9}, SimCLRv2\cite{10}, and CLIP \cite{11}. Contrastive learning achieves this by contrasting samples with their semantically similar (positive) and dissimilar (negative) counterparts. Through meticulous design of the model architecture and contrastive loss function, the method ensures that the representations of positively related pairs are drawn closer in the representation space, while the representations of negatively related pairs are pushed farther apart, thereby yielding clustering-like effects. Notably, SimCLR, introduced by Chen et al. in 2020 \cite{9}, is a prominent approach that employs contrastive learning to acquire robust visual representations devoid of manual annotation. Its primary objective is to maximize the coherence between various perspectives of the same sample, while concurrently minimizing the coherence between different samples' perspectives.

SimCLR adopts a two-step training process, including data augmentation and contrastive loss optimization. In the first step, SimCLR applies strong data augmentation techniques such as random cropping, color distortion, and Gaussian blur to create different views of each input sample. These augmented views are beneficial for learning. In the second step, SimCLR leverages a contrastive loss function to guide the model in bringing positive representations closer while simultaneously pushing negative representations further apart. The calculation of the contrastive loss hinges on the cosine similarity between the representations of the augmented views. The overarching objective is to maximize the similarity between positive pairs while minimizing the similarity between negative pairs.

The contrastive loss formula used in SimCLR is as follows:
\begin{equation}
    \ell_{i, j}=-\log \left(\frac{\exp \left(\operatorname{sim}\left(\boldsymbol{z}_{i}, \boldsymbol{z}_{j}\right) / \tau\right)}{\sum_{k=1}^{2 N} \mathbb{I}(k \neq i) \cdot \exp \left(\operatorname{sim}\left(\boldsymbol{z}_{i}, \boldsymbol{z}_{k}\right) / \tau\right)}\right)
\end{equation}
where $\mathbb{I}(k \neq i) \in\{0,1\}$ is an indicator function evaluating to 1 iff $k \neq i$. $\boldsymbol{z}_{i}$ and $\boldsymbol{z}_{j}$ represent positive pairs, $\boldsymbol{z}_{k}$ represents negative pairs, and $\tau$ is a temperature parameter that controls the sharpness of the distribution.

SimCLR has shown promising results in various computer vision tasks, including image classification, object detection, and image segmentation. Through its adeptness at acquiring comprehensive and distinctive visual representations, SimCLR has established itself as a leader in the field. This is evidenced by its attainment of state-of-the-art performance on benchmark datasets, underscoring its capacity to generate impactful and discriminative visual features.

\section{Detecting Backdoor Attack without Pre-training Dataset}
Current defense mechanisms have proven effective in mitigating or eliminating existing backdoor attacks on SSL, primarily due to their focus on static triggers. Nevertheless, the efficacy of these defense methods faces significant challenges when they are restricted from leveraging any knowledge pertaining to the pre-training dataset. Detecting and defending against such attacks solely based on information derived from downstream datasets presents a formidable challenge.

In supervised learning, STRIP \cite{strip} operates by evaluating whether overlaying the input sample onto a randomly chosen set of images leads to an increase in entropy for the prediction of class labels. If the entropy increases significantly, indicating that the class label becomes harder to predict, the input is classified as normal. Conversely, if the entropy remains low, it suggests the presence of a trigger in the input image. STRIP has demonstrated its efficacy in safeguarding against backdoor attacks characterized by patch-type trigger patterns. We directly utilize STRIP to detect backdoor samples on downstream tasks of SSL. However, its effectiveness diminishes when applied to SSL scenarios. This section presents a comprehensive analysis of the factors contributing to the reduced efficacy of the STRIP technique within the SSL scenario. To address these limitations, we propose an advanced defense mechanism named STRIP-Cl. Through empirical assessment, we showcase the defensive capabilities of both methodologies against BadEncoder \cite{1} and SSL-Backdoor \cite{saha2022backdoor}.

\textit{Defense Setting:} 
The defender's goal is to detect and mitigate backdoor samples while minimizing the impact on the overall model performance. We assume that the defender has access to the suspected encoder and obtains feature vectors of input images from the downstream dataset to train the downstream classifier. Operating under the constraints of not knowing the trigger or target categories and lacking access to reliable data or image labels, the defender faces the challenge of defending against these attacks. Our primary focus is on investigating backdoor attacks that utilize a static patch-type trigger pattern.

\subsection{STRIP and STRIP-CL } 

\subsubsection{Algorithm overview}

Assuming that the backdoor trigger is input-agnostic, the defense proposed in \cite{strip} aims to detect malicious samples before they are fed into the downstream classifier. To achieve this, the method filters suspicious images by superimposing various clean image patterns and observing the randomness of their prediction. The weight of the image superposition is equal for foreground and background images, and the randomness of the prediction is expressed through entropy. The smaller the entropy, the higher the probability that a suspicious image is malicious.


\begin{figure*}[htbp]
	\centering
	\begin{minipage}[c]{0.24\textwidth}
		\centering
		\includegraphics[width=\textwidth]{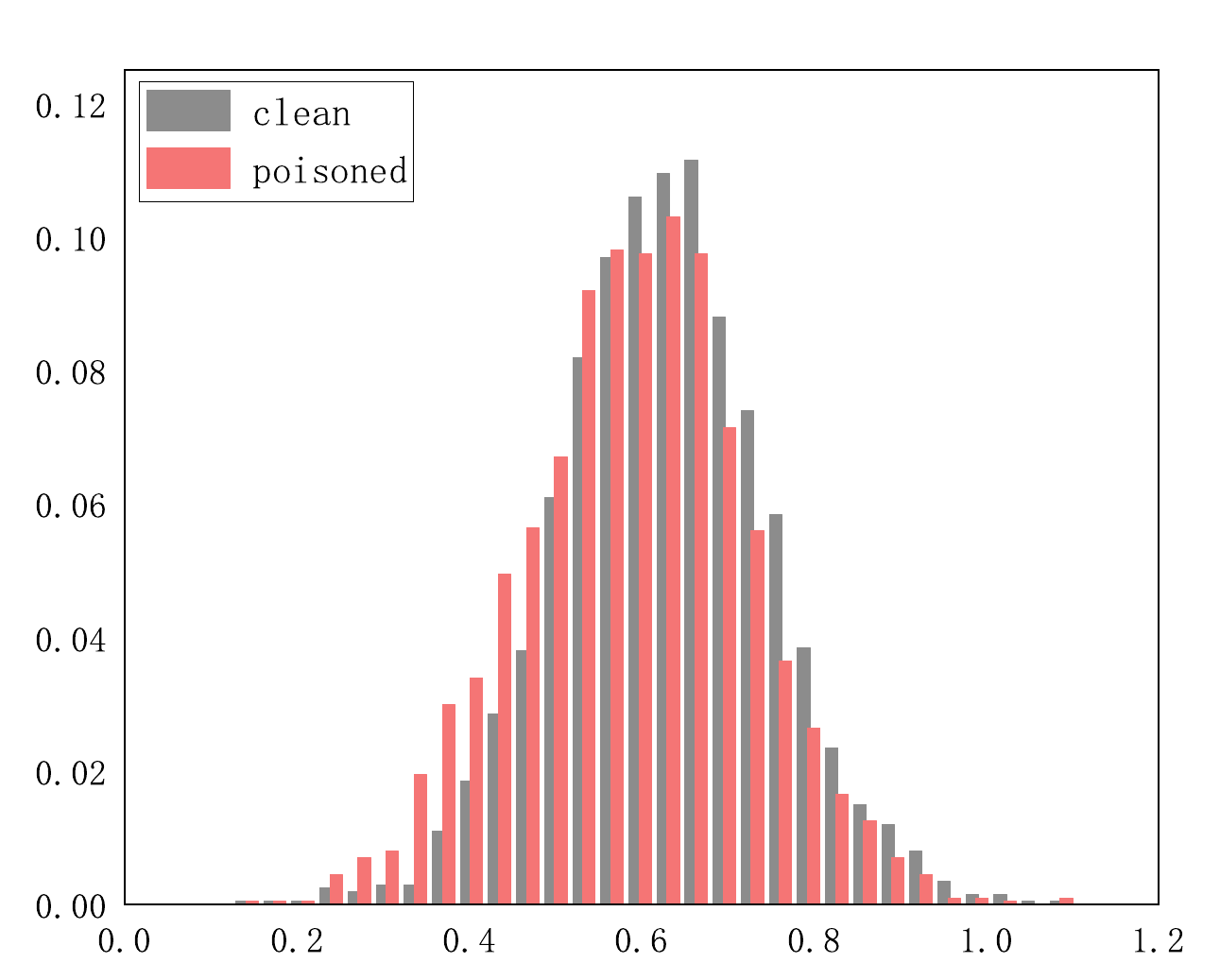}
		\subcaption{}
		\label{fig:2a}
	\end{minipage} 
	\begin{minipage}[c]{0.24\textwidth}
		\centering
		\includegraphics[width=\textwidth]{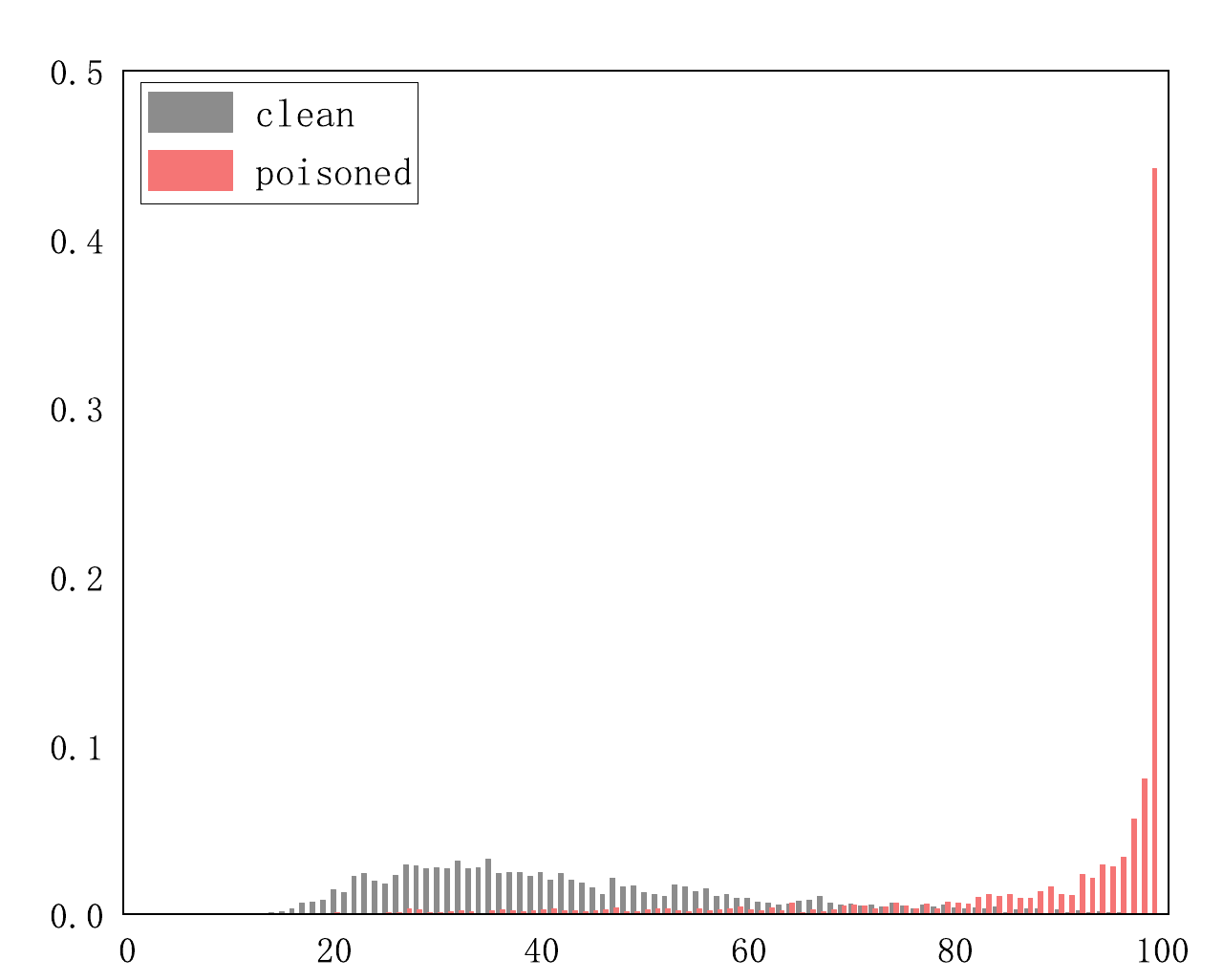}
		\subcaption{}
		\label{fig:2b}
	\end{minipage} 
	\begin{minipage}[c]{0.24\textwidth}
		\centering
		\includegraphics[width=\textwidth]{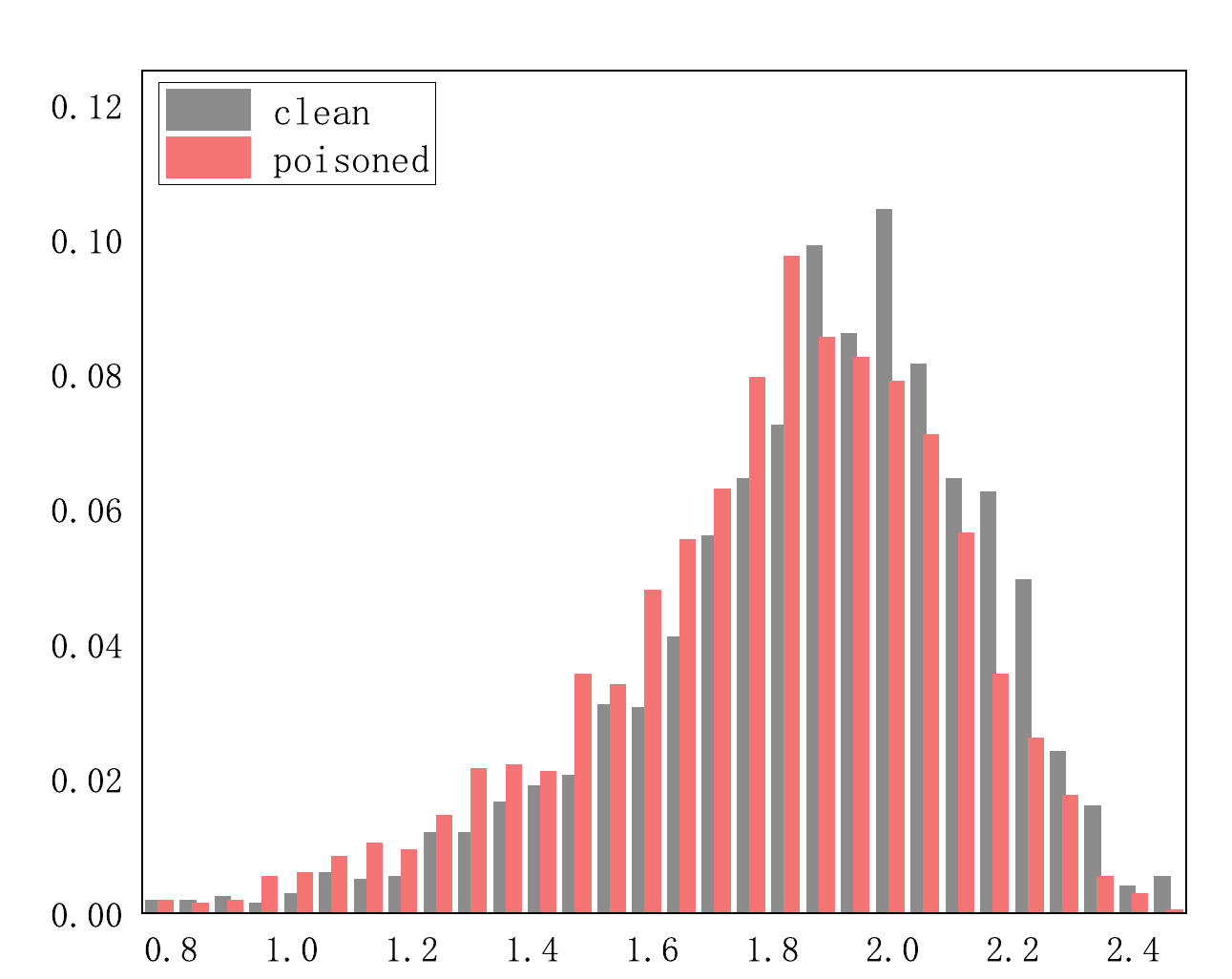}
		\subcaption{}
		\label{fig:2c}
	\end{minipage}
         \begin{minipage}[c]{0.24\textwidth}
		\centering
		\includegraphics[width=\textwidth]{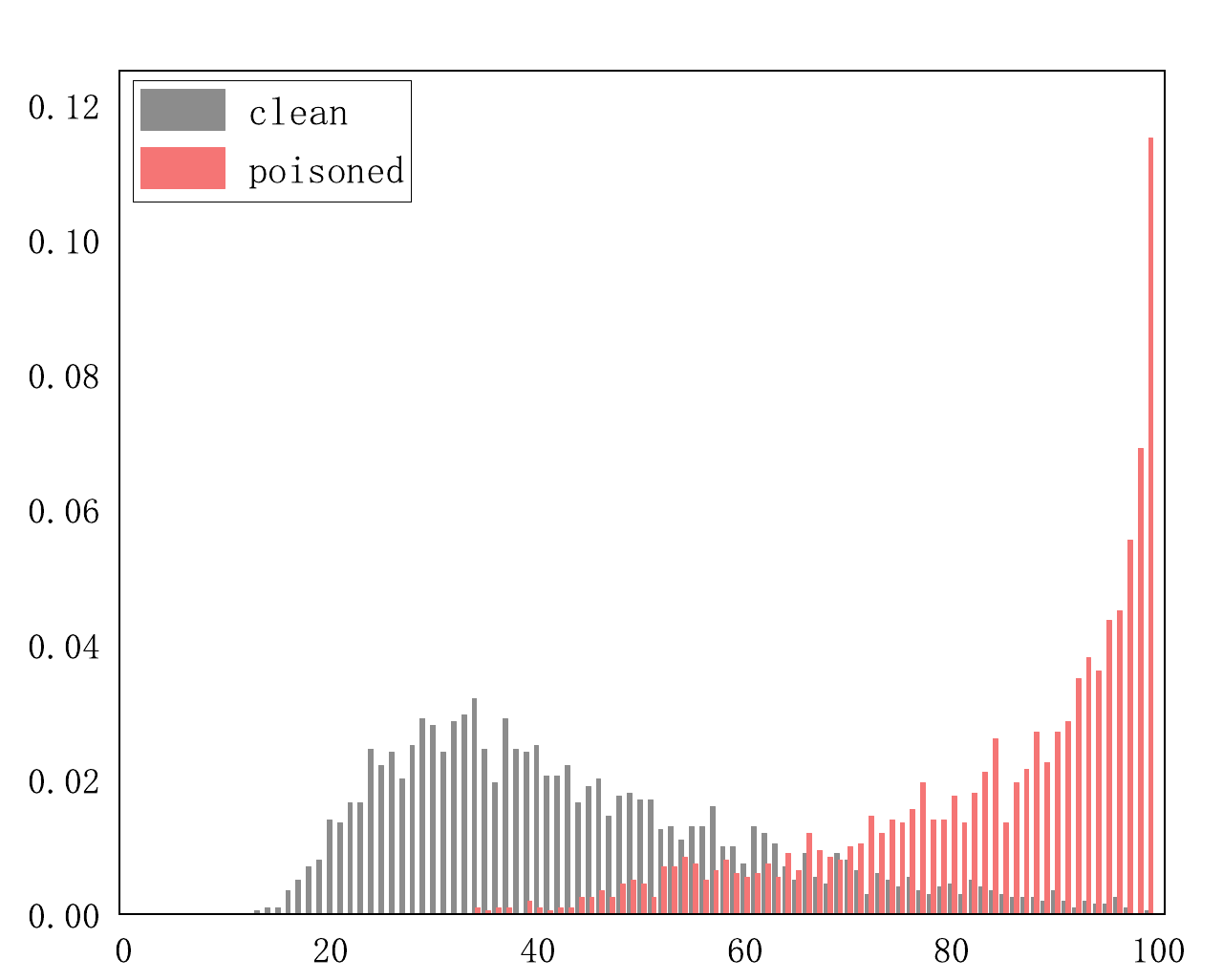}
		\subcaption{}
		\label{fig：2d}
	\end{minipage} 
	\caption{
        Maximum prediction distribution map for detecting BadEncoder and SSL-Backdoor attack, Strip and Strip-Cl.
        (a) Maximum predicted distribution of Strip defense BadEncoder; 
        (b) Maximum predicted distribution of Strip-Cl defense BadEncoder;
        (c) Maximum predicted distribution of Strip defense SSL-Backdoor; 
        (d) Maximum predicted distribution of Strip-Cl defense SSL-Backdoor.}
	\label{fig:all}
\end{figure*}

However, existing solutions designed for supervised learning face challenges when detecting backdoors in pre-trained encoders since they require classification labels. To address this limitation, we propose an enhanced version of the STRIP, called STRIP-CL, tailored specifically for contrastive learning. When using an encoder with strong feature extraction capabilities, the superimposition of image mode in STRIP has no noticeable effect on clean images. As a result, the entropy distributions of clean samples and malicious samples overlap, making them difficult to distinguish. STRIP-Cl observes the maximum prediction distribution to determine if an input image is malicious. Specifically, we generate a distribution map for STRIP-Cl using hard labels, counting the prediction outcomes from the input image classifier after superimposing multiple clean images onto each input image. The highest count predicted for a certain class serves as the abscissa ($\boldsymbol{x}$), while the ratio of samples with the maximum count of $\boldsymbol{x}$ to the total is taken as the ordinate ($\boldsymbol{y}$). We classify an image as malicious based on the abscissa threshold. In comparison to STRIP, our proposed method is better suited for detecting sample-agnostic triggers in self-supervised scenarios.

\subsubsection{Defense assessment}
Fig. \ref{fig:all} illustrates the maximum predicted distribution graphs of the STRIP and STRIP-CL defenses against BadEncoder and SSL-Backdoor. In this experiment, both BadEncoder and SSL-Backdoor utilize an upstream dataset of CIFAR10 and a downstream dataset of STL10 to construct the target model. To enhance the success rate of SSL-Backdoor, which is typically around 33.1\% \cite{zheng2023ssl}, we increase its poisoning rate to 5\%. After achieving an elevated attack success rate of 82\%, we apply the defense method.

From Fig. \ref{fig:2a} and \ref{fig:2c}, it can be observed that the original STRIP fails to defend against BadEncoder and SSL-Backdoor attacks, as the distribution of the backdoored samples generated by these two attacks overlaps with that of the clean samples, making it challenging to distinguish them through a simple threshold setting. Conversely, Fig. \ref{fig:2b} and \ref{fig：2d} demonstrate that clean samples tend to be located on the left side of the distribution map, closely following a normal distribution, while malicious samples predominantly occupy the far right side of the distribution map. Therefore, STRIP-Cl can effectively filter out and remove the backdoor samples from the backdoored training set of BadEncoder and SSL-Backdoor, significantly reducing their attack success rates. We found it necessary to adjust the weights applied in image overlay appropriately. Here, we chose a weight value of 0.9, as this setting leads to more distinct classification boundaries in the distribution map.

\subsection{Why did the STRIP defense fail } 
The core concept of STRIP \cite{strip} is to filter samples before feeding them into the model to detect the presence of a backdoor trigger. STRIP uses entropy as the basis for filtering, determining whether an image is suspicious by computing the entropy predicted by image classification. Specifically, STRIP overlays disruptions on the input image. If the input is a backdoor image, the classification outcome will be highly certain due to the trigger's presence, leading to very low entropy. Conversely, if the input is a clean image, the entropy will be relatively high. STRIP detects suspicious images by setting an entropy threshold, allowing it to differentiate between backdoor and clean images.

In supervised learning, due to the involvement of labeled information, models perform classification tasks by learning from these labels, enabling them to acquire more generalized features. The decision boundaries learned by these models tend to be clearer, ensuring that predictions for specific images do not deviate significantly from those for others. Consequently, the distribution of soft labels becomes more dispersed, and the entropy of clean images is relatively higher.

However, labeled data is absent in SSL, models must discern the relationship between triggers and samples to execute classification tasks. This can lead to fuzzier decision boundaries, less dispersed distributions of soft labels, and consequently, lower entropy for clean samples. This results in indistinct threshold boundaries in the entropy distribution plots between backdoor and clean images. In such scenarios, the filtering mechanism of STRIP may mistakenly regard images as non-suspicious, leading to inaccurate detection of backdoor samples. Therefore, in the context of SSL, the defensive performance of STRIP significantly diminishes.

In contrast, our proposed detection framework, STRIP-Cl, focuses on the distribution of hard labels. Instead of using entropy distribution, we employ the maximum predicted distribution of hard labels for sample detection. Entropy has minimal influence on hard labels, allowing for a more precise reflection of the classifier's judgment concerning input images. Consequently, STRIP-Cl enhances the defensive capabilities of STRIP, broadens its defense scope, and enhances model security.

\section{Methodology}
Our proposed attack as follows. The definition of the notations is presented in Table \ref{tab:notation}.

\begin{table*}[]
    \centering
    \caption{A Summary of Notations.}
    \resizebox{1.0\columnwidth}{!}{
    \begin{tabular}{ccc}
       \toprule[1.2pt]
       Name & Notation & Meaning  \\
       \midrule
       Clean Encoder  & $e$ & The target encoder used to inject the backdoor \\
       Backdoored Encoder  & $e^{\prime}$ & An encoder that has been injected with a backdoor \\
       (Target downstream task, target class)  & $(T_i, c_i)$ & A pair of target downstream tasks and target classes \\
       Reference Inputs  & $R_i$ & one or more images in the target class for each (target downstream task, target class) pair \\
       A Normal Sample &  $\boldsymbol{x}$ & A clean input image   \\
       A Backdoor Trigger  & $\boldsymbol{p_x}$ & A disturbance that triggers a back door \\
       A Backdoor Sample & $\boldsymbol{x} \oplus \boldsymbol{p}_{x}$  &  A backdoor instance as the model input   \\
       Shadow Dataset  & $\mathcal{D}_{s}$ & A collection of backdoor images \\
       Hyper-parameter  & $\lambda_{1}$ & Parameters used by the backdoor injection process \\
       Steganographic encoder  & $e_i$ &  An encoder that adds a watermark to a clean image \\
       \bottomrule[1.2pt]
    \end{tabular}}
    \label{tab:notation}
\end{table*}

\subsection{Threat Model}\label{AA}
We only consider backdoor attacks on visual encoders. The attacker's objective is to inject a backdoor into a pre-trained image encoder, enabling downstream classifiers constructed based on the backdoored image encoder to predict the attacker's chosen prediction for inputs containing the attacker's chosen trigger.

\textbf{Attacker’s Capacities.} Our work considers two possible attackers: 1) Untrusted service providers who may inject backdoors into their pre-trained image encoders and provide them to downstream users, 2) Malicious third parties who inject backdoors into image encoders pre-trained by service providers and publish them on the internet for downstream users to download and use. The attacker has access to the watermark dataset used for training the watermark generator, clean pre-trained image encoders, unlabelled images (referred to as shadow dataset), and some images for each pair (target downstream task, target class) known as reference inputs. However, the attacker cannot access the downstream dataset used to construct the downstream classifiers or influence the training process of the downstream classifiers.

\textbf{Attacker’s Goals.} In general, backdoor attackers intend to embed hidden backdoors in DNNs through data poisoning. The hidden backdoor will be activated by the attacker-specified trigger, i.e., the prediction of the image containing the trigger will be the target label, no matter what its ground-truth label is. In particular, attackers have three main goals, including effectiveness, stealthiness, and sustainability. The effectiveness requires that the prediction of attacked DNNs should be the target label when the backdoor trigger appears, and the performance on benign testing samples will not be significantly reduced. The stealthiness requires that adopted triggers should be concealed and invisible to the human eye. The sustainability requires that the attack should still be effective under some common backdoor defenses.

\subsection{The Proposed Attack }

In this section, we will elucidate our proposed method, GhostEncoder. Before delving into the generation of sample-specific triggers, we provide a concise overview of the GhostEncoder procedure. Our approach is formulated as the following optimization problem.

Formally, to achieve the goals of effectiveness and practicality mentioned earlier, we propose modifying the clean image encoder $e$ to obtain a backdoor encoder $e^{'}$. For each (target downstream task, target class) pair $(T_i, C_i)$, the attacker selects a set of reference inputs $\mathcal{R}_{i}=\left\{\boldsymbol{x}_{i 1}, \boldsymbol{x}_{i 2}, \cdots, \boldsymbol{x}_{i r_{i}}\right\}$ from the target class $C_i$, where $\boldsymbol{p}_{x}$ is the trigger generated for each input $\boldsymbol{x}$. $\boldsymbol{x}\oplus \boldsymbol{p}_{x}$ is the backdoor input. We propose using efficiency loss and utility loss to quantify the efficiency and utility objectives, respectively. Our effectiveness loss consists of the following two formulas:

\begin{equation}
    L_{0}=-\frac{\sum_{i=1}^{t} \sum_{j=1}^{r_{i}} \sum_{\boldsymbol{x} \in \mathcal{D}_{s}} s\left(e^{\prime}\left(\boldsymbol{x} \oplus \boldsymbol{p}_{x}\right), e^{\prime}\left(\boldsymbol{x}_{i j}\right)\right)}{\left|\mathcal{D}_{s}\right| \cdot \sum_{i=1}^{t} r_{i}}
\end{equation}

\begin{equation}
    L_{1}=-\frac{\sum_{i=1}^{t} \sum_{j=1}^{r_{i}} s\left(e^{\prime}\left(\boldsymbol{x}_{i j}\right), e\left(\boldsymbol{x}_{i j}\right)\right)}{\sum_{i=1}^{t} r_{i}}
\end{equation}
where $s(,)$ measures the similarity between two feature vectors, $|\mathcal{D}_{s}|$
represents the number of inputs in the shadow dataset, Our effectiveness loss is a weighted sum of the two terms $L_{0}+\lambda_{1} \cdot L_{1}$, where $\lambda_{1}$ is a hyperparameter to balance the two terms.

The effectiveness loss achieves two goals: 1)The backdoor encoder $e^{\prime}$ generates similar feature vectors for the reference input and the backdoor input embedded in the trigger in the shadow data set. 2) The backdoor image encoder $e^{\prime}$ and the clean image encoder $e$ generate similar feature vectors for the reference input.

The main goal of the utility loss is to preserve the accuracy of downstream classifiers that rely on our backdoored image encoder for clean inputs. This requires ensuring that the backdoor image encoder and the clean image encoder generate similar feature vectors for clean inputs. To achieve this goal, we define a utility loss :

\begin{equation}
    L_{2}=-\frac{1}{\left|\mathcal{D}_{s}\right|} \cdot \sum_{\boldsymbol{x} \in \mathcal{D}_{s}} s\left(e^{\prime}(\boldsymbol{x}), e(\boldsymbol{x})\right)
\end{equation}

After defining the three loss terms $L0$, $L1$ and $L2$, we tackle the optimization problem of the backdoor image encoder $e^{\prime}$ based on these terms.

\begin{equation}
    \min _{e^{\prime}} L=L_{0}+\lambda_{1} \cdot L_{1}+\lambda_{2} \cdot L_{2}
\end{equation}

The two hyperparameters $\lambda_{1}$, $\lambda_{2}$ are employed to balance these loss terms. Their impact on the backdoor encoder will be thoroughly examined during the evaluation process.

\begin{figure*}[!h]
    \centering
    \includegraphics[width=1.0\linewidth]{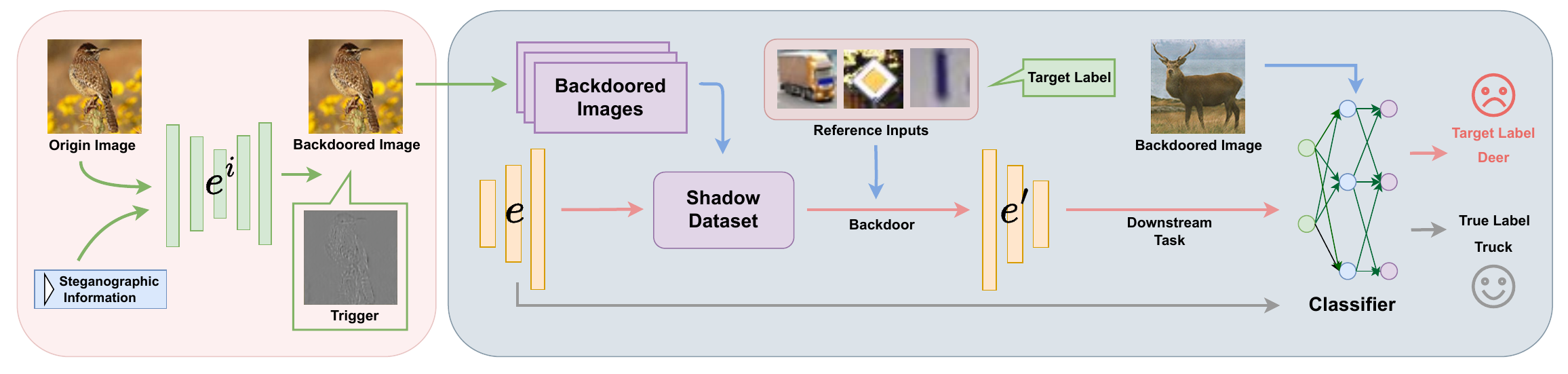}
    \caption{An Overview of proposed GhostEncoder.}
    \label{fig:main}
\end{figure*}

\textbf{Pipline:} In this work, we propose a novel dynamic invisible backdoor attack in the context of self-supervised learning, named GhostEncoder. Fig. \ref{fig:main} provides an overview of GhostEncoder. The left part illustrates the generation stage of backdoored images. Here, we employ an additional watermark encoder, denoted as $e^i$, to embed hidden information into the backdoor images, generating the triggers. This encoder is derived from a pre-trained encoder-decoder network using a watermark dataset. On the right side of Fig. \ref{fig:main}, we present the stages of backdoor injection and downstream task inference. The obtained embedded triggers in the backdoor images serve as the shadow dataset, while the reference inputs consist of images collected from the target class on the web. The objective of the backdoor injection phase is to construct a backdoor image encoder $e^{\prime}$ from a clean image encoder $e$, achieving the goals of effectiveness, invisibility, and robustness. During the inference stage, the downstream classifier built upon our backdoor image encoder will predict any input containing the corresponding trigger to be of the target class, without affecting the normal performance of non-target downstream tasks.

\subsection{How to Generate Trigger}
Inspired by DNN-based image steganography, ISSBA\cite{2} encodes a specified string into benign images as backdoor images using a pre-trained encoder. However, this method operates in a supervised scenario, requiring more annotated data and utilizing target label information as the content of the steganographic string. In contrast, our method is implemented in a self-supervised setting, freeing itself from label constraints and not limiting the specific content of the string. It only requires consistent encoding information in the backdoored images.

\begin{figure*}[t]
    \centering
    \includegraphics[width=1.0\columnwidth]{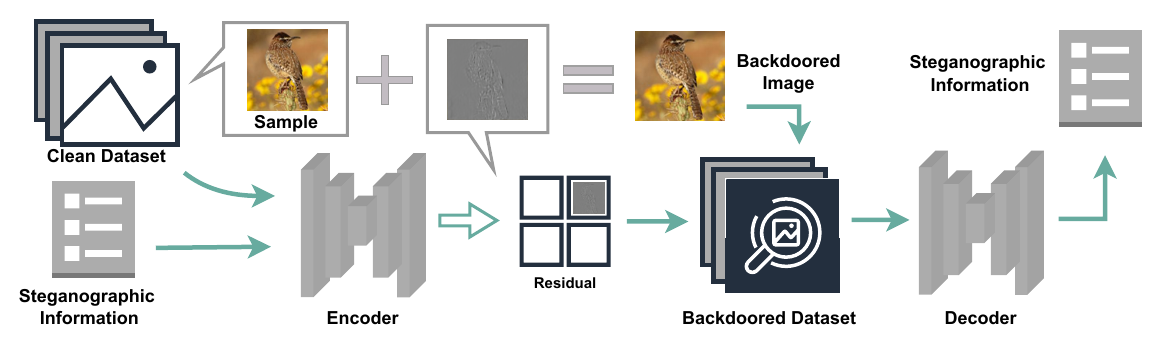}
    \caption{The training process of encoder-decoder network.}
    \label{fig:encoder}
\end{figure*}

\textbf{Generating Sample-Specific Triggers.}
We exemplify with a pre-trained encoder-decoder network to generate sample-specific triggers. The watermark encoder's output varies based on the input image, yielding invisible additional noise as sample-specific triggers. As depicted in Fig. \ref{fig:encoder}, we create the backdoored dataset by adding watermarks to the clean dataset. The encoder first takes a benign image and watermark information to generate a residual image (i.e., the corresponding trigger). Overlaying this residual image onto the benign image yields the encoded image. The watermark encoder and decoder are jointly trained on the benign training set (watermark dataset). Throughout the training, a uniform string functions as steganographic content, training the watermark encoder to embed it into images, simultaneously minimizing the perceptual difference between input and encoded images. The decoder, on the other hand, learns to recover hidden messages from encoded images. Using this well-trained watermark encoder, we append sample-specific triggers to input images, generating backdoor images.

\section{Evaluation}
\subsection{Experimental Setup}\label{AA}

\subsubsection{Datasets}

In this work, we employ four distinct image datasets, namely CIFAR10, STL10, GTSRB, and SVHN, to facilitate comprehensive evaluation.
 \begin{itemize}
   \item \textbf{CIFAR10}: This dataset comprises 50,000 training images and 10,000 testing images, with each image sized at 32x32x3 pixels. The images are categorized into ten different classes.
   \item \textbf{STL10}: Consisting of 5,000 labeled training images and 8,000 labeled testing images, the STL10 dataset includes images sized at 96x96x3 pixels. It encompasses ten distinct classes, and notably, there are an additional 100,000 unlabeled images available.
   \item \textbf{GTSRB}: Comprising 51,800 traffic sign images distributed across 43 categories, the GTSRB dataset is divided into 39,200 training images and 12,600 testing images.
   \item \textbf{SVHN}: With each image representing a digit from Google Street View's house number dataset, the SVHN dataset contains 73,257 training images and 26,032 testing images, all sized at 32x32x3 pixels. Additionally, the dataset is characterized by some scattered digits near the digits of interest, rendering it a noisy dataset.
\end{itemize}

\subsubsection{Parameter setting}

In our experiments, we employed CIFAR-10 or STL-10 as the pre-training datasets due to their larger image capacity and non-noisy nature. For both of these pre-training datasets, we selected the ResNet-18\cite{he2016deep} architecture as the framework for our image encoders. During training, we utilized SimCLR\cite{9} and applied a collection of augmentation operators, including random resizing-cropping, horizontal flipping, color distortion, and random grayscale. We employed the Adam optimizer and conducted training for 1000 epochs to obtain the image encoders. The implementation of this process was based on publicly available code from SimCLR\footnote{https://github.com/google-research/simclr}\footnote{https://github.com/leftthomas/ SimCLR}.

The triggers used in our approach are generated by watermark encoders trained on watermark datasets. The watermark encoder is trained using an encoder-decoder network configuration consistent with StegaStamp \cite{tancik2020stegastamp}. In this configuration, the encoder network adopts a U-Net\cite{ronneberger2015u} style, and the decoder network employs a spatial transformer network\cite{jaderberg2015spatial}. The training incorporates four loss terms: L2 residual regularization, LPIPS perceptual loss\cite{zhang2018unreasonable} to minimize perceptual distortion, cross-entropy loss for concealed information reconstruction, and perceptual distortion loss. Scaling factors for these losses are set to 2.0, 1.5, 0.5, and 1.5, respectively. During training, we employed both the Adam\cite{kingma2014adam} optimizer and the SGD optimizer, with initial learning rates set to 0.0001 and 0.001, respectively. The learning rate is decayed by a factor of 0.1 after the 15th and 20th epochs. The resulting encoder is utilized to create backdoor images for generating a shadow dataset, consisting of 50,000 images extracted from the pre-training dataset.

The Backdoor injection phase begins with the assumption that the attacker selects a target downstream task/dataset and a target category. When the upstream dataset is CIFAR-10, we employ STL-10, GTSRB, and SVHN as downstream datasets. In the case of an upstream dataset of STL-10, we opt for CIFAR-10, GTSRB, and SVHN as downstream datasets. Additionally, "airplane," "truck," "priority sign," and "digit one" are chosen as target categories for CIFAR-10, STL-10, GTSRB, and SVHN datasets, respectively. We assume a reference input for each (target downstream task, target category) pair, collected by correctly classifying acquired reference inputs as the target category using the backdoored downstream classifier. We perform fine-tuning on the pre-trained image encoder with a learning rate of 0.001 and a batch size of 256 for 200 epochs to inject the backdoor unless otherwise specified. Parameters $\lambda_{1}$ and $\lambda_{2}$ default to 1 unless specified otherwise.

Given an image encoder (with a backdoor) pre-trained using a pre-training dataset, we utilize it as the feature extractor for downstream tasks. For a chosen dataset as the downstream dataset, its training data train the downstream classifier, while its testing data evaluate the classifier. Specifically, we employ a two-layer Multilayer Perceptron (MLP) with a hidden layer size of 128 as the downstream classifier, employing the cross-entropy loss function and Adam optimizer during training. Furthermore, the downstream classifier is trained for 500 epochs with an initial learning rate of 0.0001. Unless stated otherwise, CIFAR-10 serves as the default pre-training dataset, and STL-10 is the default downstream dataset. Note that we resize each image in the STL-10 dataset to a size of 32×32×3 to maintain consistency with other datasets.


\subsubsection{Evaluation metrics} 
In order to effectively assess the performance of the attacks, we primarily employ three metrics, namely Clean Accuracy (CA), Backdoored Accuracy (BA), and Attack Success Rate (ASR), to evaluate our BadEncoder targeting the classification model. Here, CA and BA represent the classification accuracy of downstream classifiers constructed based on clean and backdoored encoders, respectively, on the downstream clean test dataset. ASR indicates the proportion of trigger-embedded samples in which the downstream classifier built upon the backdoored encoder predicts the corresponding target class.

Furthermore, we introduce two additional metrics for supplementary evaluation: ASR-B and Attack Success Rate after Defense (ASR-AD). ASR-B measures the percentage of trigger-embedded backdoor images recognized as the target class by the downstream classifier constructed with a clean encoder, reflecting the success rate of attacks when the attacker has not injected a backdoor into the pre-trained image encoder.
To assess the robustness of the defense method against attacks, we introduce the ASR-AD. It quantifies the proportion of test inputs with embedded triggers accurately predicted as the target class by the backdoored downstream classifier after undergoing the Strip-Cl defense strategy.

\subsection{Effectiveness and Efficiency of Attack}

\begin{table*}[!t]
\centering
\caption{Attack effect comparison between BadEncoder and GhostEncoder.} \label{tab:aStrangeTable}
\resizebox{1.0\columnwidth}{!}{
\begin{tabular}{c|c|c|cc|ccc}
\toprule[1.2pt]
Pre-training Dataset& Method& Downstream Dataset& CA (\%)& BA (\%)& ASR-B (\%)& ASR (\%)& ASR-after Defense(\%) \\ \cmidrule(r){1-8}
\multirow{6}{*}{CIFAR10} & \multirow{3}{*}{BadEncoder} & STL10 & 76.14 & 76.18 & 10.37 & 99.73 & 15.8 (83.93 ↓)\\
&&GTSRB&81.84&82.27&5.09&98.64&14.5 (84.14 ↓)\\ 
&&SVHN&61.72&69.32&35.17&99.14&3.2 (95.94 ↓)\\  \cmidrule(r){2-8}
&\multirow{3}{*}{GhostEncoder}&STL10&76.14&76.28&8.61&96.57&\textbf{95.30} (1.27 ↓)\\
&&GTSRB&81.84&64.92&4.83&65.19&\textbf{65.05} (0.14 ↓)\\
&&SVHN&61.72&55.98&20.98&24.41&\textbf{24.05} (0.36 ↓)\\
\midrule
\multirow{6}{*}{STL10} & \multirow{3}{*}{BadEncoder} & CIFAR10 & 87.36 & 86.82 & 10.13 & 94.35 & 21.0 (73.35 ↓)\\
&&GTSRB&76.83&78.34&4.86&88.36&19.88 (68.48 ↓)\\ 
&&SVHN&54.72&62.91&34.48&23.37&21.45 (1.92 ↓)\\ \cmidrule(r){2-8}
&\multirow{3}{*}{GhostEncoder}&CIFAR10&87.36&84.97&10.02&90.28&\textbf{87.82} (2.46 ↓)\\
&&GTSRB&76.83&70.54&3.68&67.16&\textbf{66.26} (0.90 ↓)\\
&&SVHN&54.72&50.62&19.18&27.82&\textbf{26.44} (1.38 ↓)\\
\bottomrule[1.2pt]
\end{tabular}
}
\label{tab:main_table}
\end{table*}


As shown in Table \ref{tab:main_table}, we evaluated BadEncoder and GhostEncoder on three different downstream datasets, each with two pretraining datasets. We provided the CA and ASR-B for the pre-trained image encoders without injecting backdoors, as well as the BA, ASR, and ASR-AD for GhostEncoder. Overall, our GhostEncoder achieved the following:

\textbf{High attack success rate without compromising model utility:}
Through experimentation, we discovered that the effectiveness of downstream attacks is influenced by the strength of the watermark. Depending on the distribution of the downstream dataset, the closer the distribution of the downstream dataset is to the pre-training dataset, the stronger the watermark is required. Consequently, we trained a robust watermark encoder for STL-10 and a less potent watermark encoder for GTSRB and SVHN, respectively.

Based on the results obtained from using two different pretraining datasets on three different downstream datasets, GhostEncoder demonstrated a high ASR. For instance, when the pretraining dataset was CIFAR10 and the downstream dataset was STL10, GhostEncoder achieved an ASR of 96.57\%. In contrast, without injecting a backdoor into the pre-trained image encoder, the ASR-B was only 8.61\%. The decrease in BA relative to CA was only 0.14\% (CA: 76.14\%, BA: 76.28\%). Thus, our GhostEncoder maintained the accuracy of the downstream classifier while achieving a high ASR. We noted that the ASR-B was relatively higher when the downstream dataset was SVHN due to its class imbalance, with the selected target class being the most popular category.

\textbf{Resistance against state-of-the-art defenses:}
ASR-AD represents the results of applying the STRIP-Cl method (introduced in Section 3.1) for defense. It involves setting a threshold to filter out toxic datasets, resulting in the filtered ASR-AD. As described in Section 3.1, the threshold was determined by selecting the maximum count of hard-label predictions for a specific class as the x-coordinate and setting a threshold to filter out toxic samples. According to the results in the ASR-AD column of the table, values in parentheses indicate the threshold. These results demonstrate that our GhostEncoder can resist STRIP-Cl defense. For example, when the pretraining dataset was CIFAR10 and the downstream dataset was GTSRB, GhostEncoder achieved a defense success rate of 65.05\%, compared to an ASR of 65.19\% without defense, resulting in a mere 0.14\% decrease in ASR. Importantly, in most cases, the difference between the ASR and the defense success rate was within 0.5\%. Thus, STRIP-Cl had minimal impact on weakening GhostEncoder's attack effectiveness. In contrast, the ASR of BadEncoder significantly decreased after the defense. For instance, when the pretraining dataset was CIFAR10 and the downstream dataset was GTSRB, its ASR dropped from 98.64\% to 14.5\%, rendering the attack nearly ineffective. In most cases, STRIP-Cl was able to completely defend against BadEncoder.

\subsection{Hyperparameter Analysis}

\subsubsection{Impact of shadow dataset}

\begin{figure*}[htbp]
	\centering
	\begin{minipage}[c]{0.30\textwidth}
		\centering
		\includegraphics[width=\textwidth]{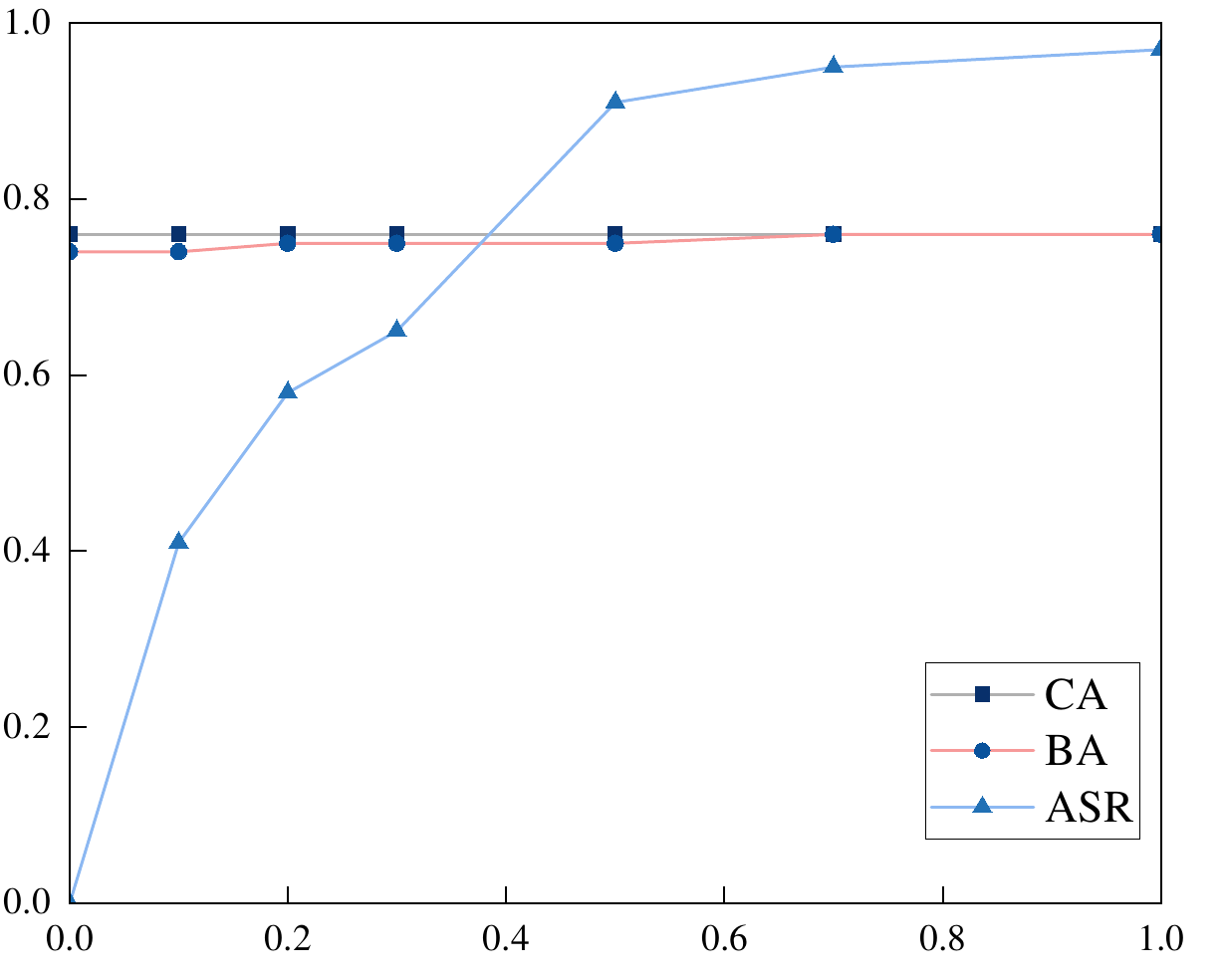}
		\subcaption{}
		\label{fig:5a}
	\end{minipage} 
	\begin{minipage}[c]{0.30\textwidth}
		\centering
		\includegraphics[width=\textwidth]{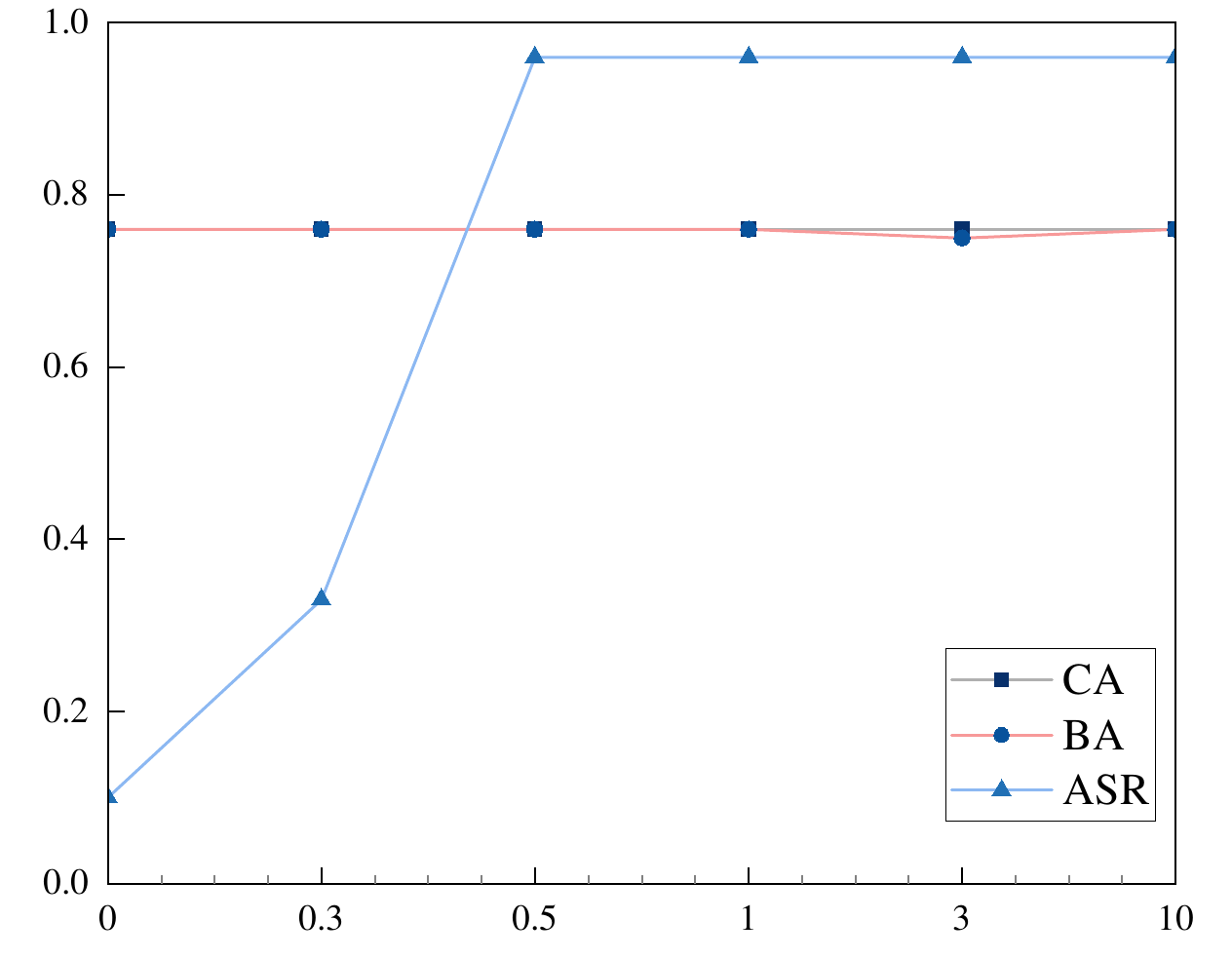}
		\subcaption{}
		\label{fig:5b}
	\end{minipage} 
	\begin{minipage}[c]{0.30\textwidth}
		\centering
		\includegraphics[width=\textwidth]{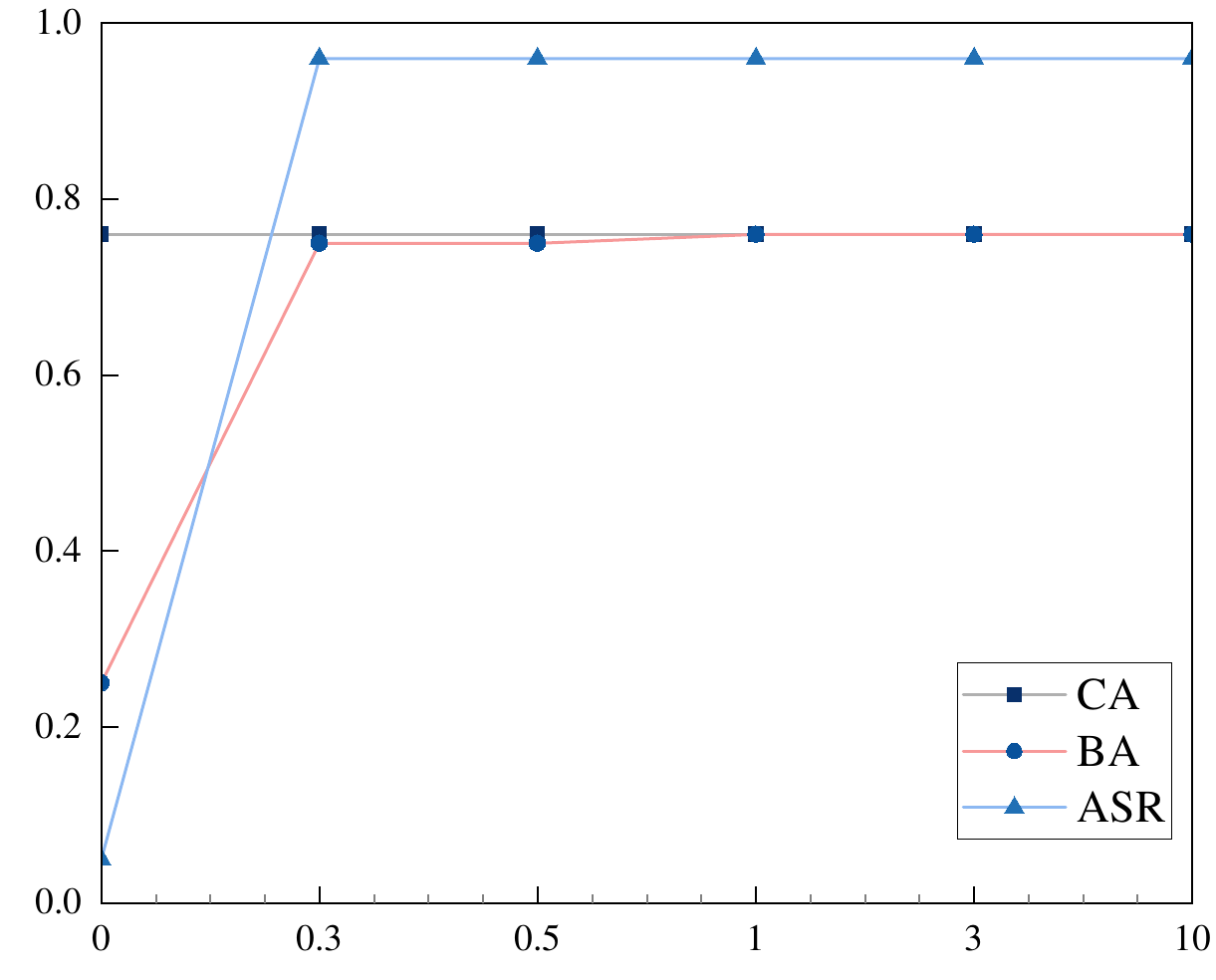}
		\subcaption{}
		\label{fig:5c}
	\end{minipage}
	\caption{
        The impact of the various parameters on BadEncoder when the pre-training dataset is cifar10 and the target downstream dataset is STL10.
    (a)The impact of the shadow dataset size
    (b)The impacts of $\lambda_{1}$ 
    (c)The impacts of $\lambda_{2}$.}
	\label{fig:param}
\end{figure*}

The shadow dataset can be characterized by its size and distribution. Therefore, we investigate the impact of the shadow dataset size on GhostEncoder. Fig. \ref{fig:5a} illustrates the influence of shadow dataset size on GhostEncoder. Taking STL10 as the target downstream dataset, we observe that when the shadow dataset size is around 25\% of the pre-training dataset, GhostEncoder achieves an ASR of over 60\%. Once the shadow dataset size surpasses about 50\% of the pre-training dataset, GhostEncoder's ASR reaches as high as 96\%, nearly approaching the performance achieved when the shadow dataset size equals the pre-training dataset, while maintaining the accuracy of the downstream classifier. This suggests that our injected dataset for the backdoor requires only a minimal amount of information from the pre-training dataset.

\subsubsection{Impact of other parameters}

Our GhostEncoder utilizes a loss function described by Equation 5, which consists of three loss terms: $L_{0}$, $L_{1}$, and $L_{2}$. We use two hyperparameters, $\lambda_{1}$ and $\lambda_{2}$, to compute the weighted loss. $L_{1}$ serves the effectiveness objective, achieving a high success rate in backdoor attacks, while $L_{2}$ serves the utility objective, maintaining high accuracy after significant perturbation, thus preserving the accuracy of the downstream classifier. By controlling the parameters $\lambda_{1}$ and $\lambda_{2}$, we can balance the accuracy of perturbation and the success rate of the attack. Therefore, we investigated the influence of $\lambda_{1}$ and $\lambda_{2}$ on our GhostEncoder. Fig. \ref{fig:5b} ($\lambda_{2}$=1) and Fig. \ref{fig:5c} ($\lambda_{1}$=1) illustrate the results.

Initially, we observe that GhostEncoder achieves high attack success rates and maintains classification accuracy when $\lambda_{1}$ and $\lambda_{2}$ exceed certain thresholds. For instance, when $\lambda_{1}$ is set to 0.5, GhostEncoder already achieves a high ASR of 96.18\%. Furthermore, GhostEncoder exhibits higher sensitivity to smaller numerical ranges of $\lambda_{1}$, while for larger numerical values of $\lambda_{1}$, even when reaching 10, the attack success rate remains stable. This stability arises from the relatively small training set size of the STL10 dataset. The distribution of $\lambda_{2}$ is similar to that of $\lambda_{1}$, showing heightened sensitivity to smaller numerical ranges and maintaining a consistent ASR without fluctuations or declines in larger numerical ranges. We attribute this behavior to the shadow dataset's size (related to L2), which significantly surpasses the quantity of reference inputs.

\begin{figure*}[!th]
    \centering
    \includegraphics[width=1.0\columnwidth]{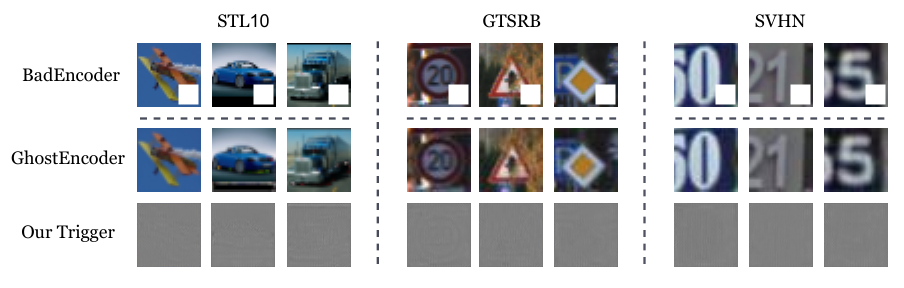}
    \caption{Comparison of the backdoor samples of BadEncoder and GhostEncoder. The third line is the trigger generated by the GhostEncoder attack.
     }
    \label{fig:hidden}
\end{figure*}

\subsection{Attack Stealthiness}

Fig. \ref{fig:hidden} presents the backdoored images generated using the BadEncoder and our GhostEncoder attack methods. Here, we showcase the backdoored images for three downstream datasets: STL10, GTSRB, and SVHN, along with the trigger images utilized in our proposed GhostEncoder method. The BadEncoder's trigger consists of a fixed 10×10 white square, positioned at the bottom right corner of the image. This trigger is singular, fixed, and clearly visible, lacking effective concealment. In contrast, the trigger employed in the backdoored images of our GhostEncoder method is sample-specific, exhibiting a remarkably high level of concealment, and imperceptible to the human eye.

\subsection{Robustness to Defense}

In this work, we investigate two emerging defenses for SSL: our proposed Strip-Cl and SSL-Cleanse. Through experiments, we demonstrate that our GhostEncoder attack can successfully bypass both of these defenses.

\subsubsection{Defense of Strip-Cl}

\begin{figure*}[htbp]
	\centering
	\begin{minipage}[c]{0.35\textwidth}
		\centering
		\includegraphics[width=\textwidth]{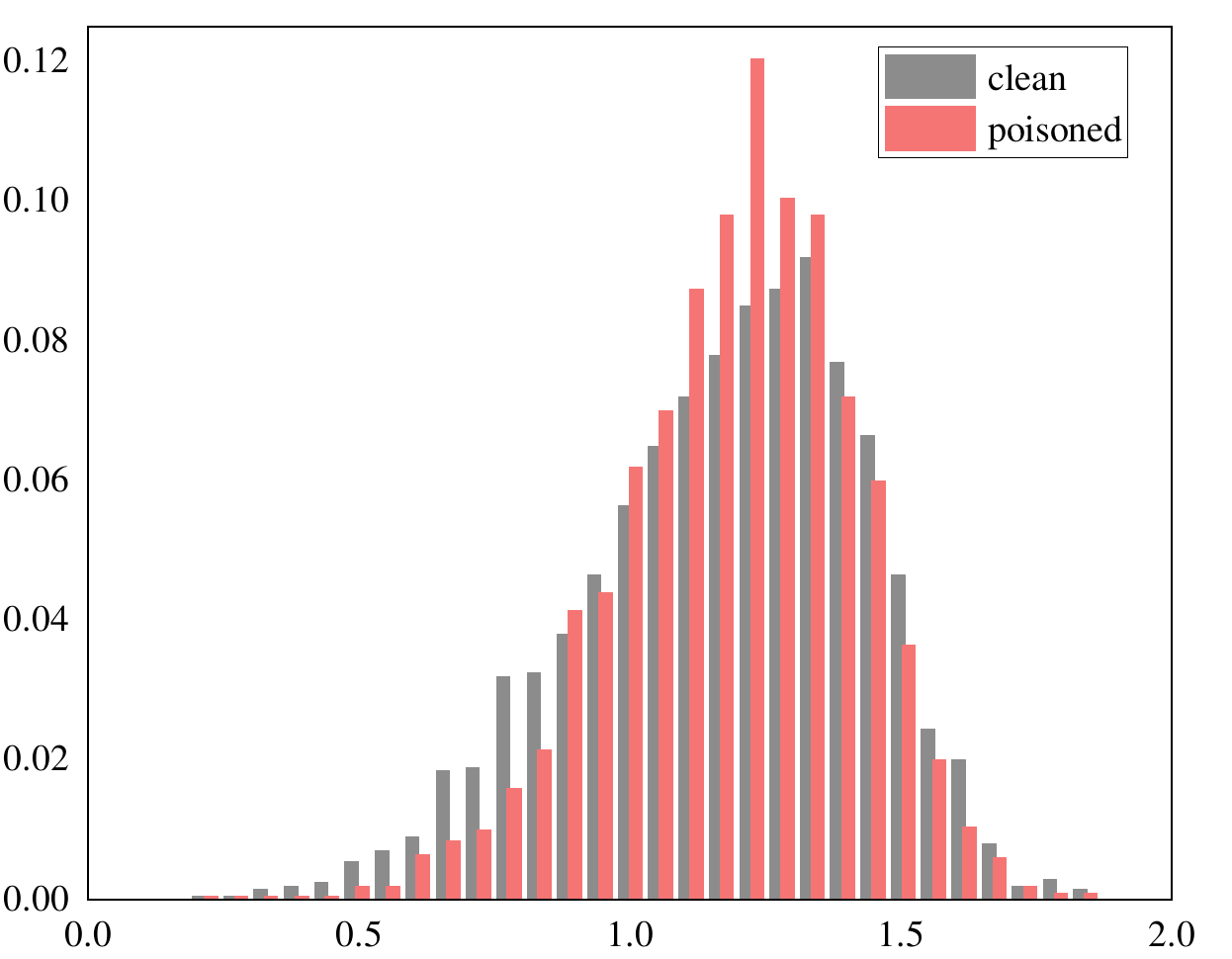}
		\subcaption{}
		\label{fig:7a}
	\end{minipage} 
	\begin{minipage}[c]{0.35\textwidth}
		\centering
		\includegraphics[width=\textwidth]{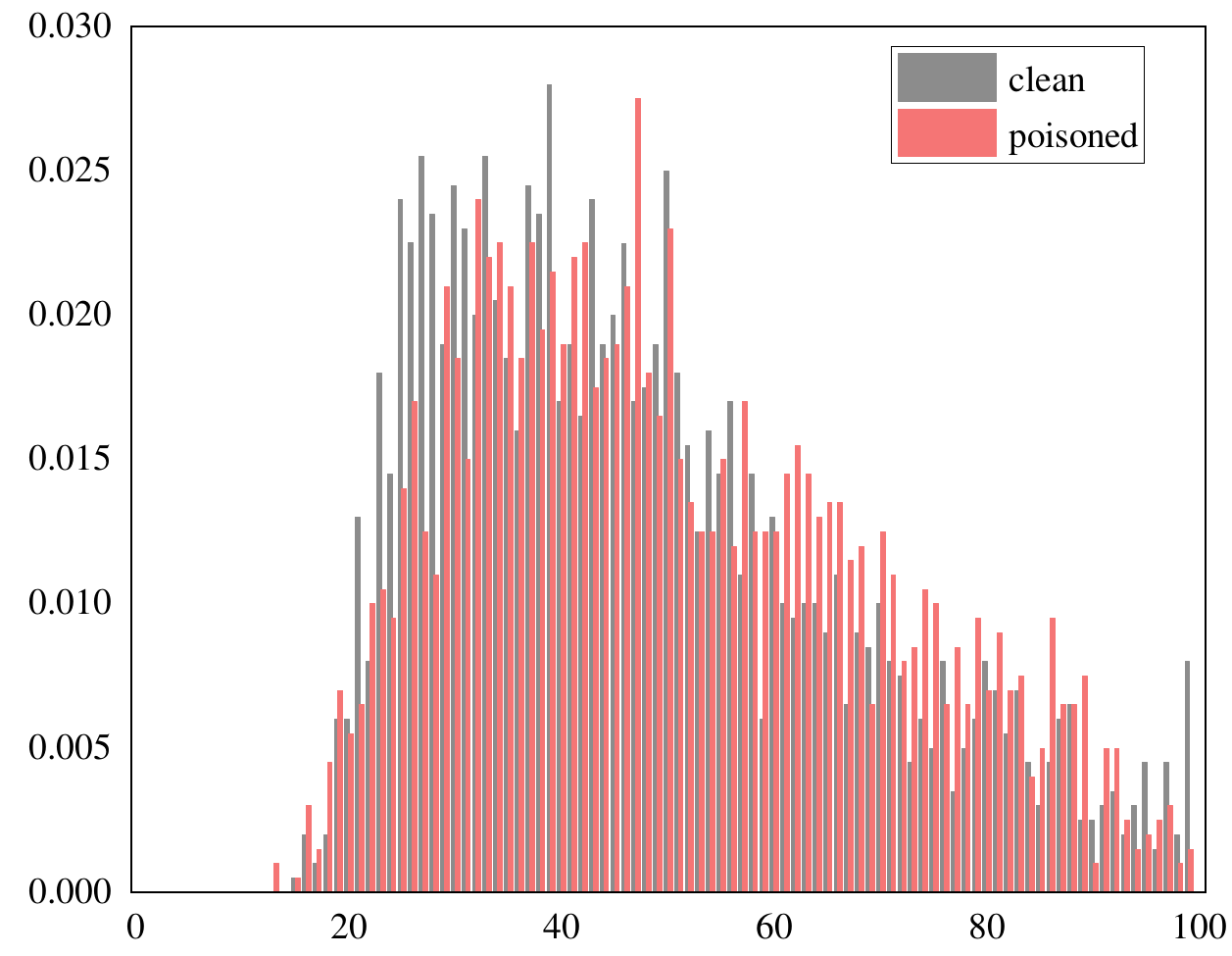}
		\subcaption{}
		\label{fig:7b}
	\end{minipage} 
	\caption{
        Maximum prediction distribution map for detecting GhostEncoder, Strip and Strip-Cl.
        (a) Maximum predicted distribution of Strip defense GhostEncoder; 
        (b) Maximum predicted distribution of Strip-Cl defense GhostEncoder.}
	\label{fig:stega}
\end{figure*}

As shown in Fig. \ref{fig:stega}, Fig. \ref{fig:7a} and \ref{fig:7b} illustrate the maximum prediction distribution maps obtained by the original STRIP method and our proposed STRIP-Cl method for defending against GhostEncoder with hard labels. From the figure, it can be observed that regardless of using STRIP or STRIP-Cl defense, the distributions of backdoored and clean samples generated by GhostEncoder are highly overlapping. Distinguishing backdoored samples from clean ones through simple thresholding becomes challenging. Therefore, both methods fail to defend against GhostEncoder effectively.


Regarding patch-based perturbation approaches, including data poisoning and model weight modification types of backdoor attacks, one common characteristic is the use of a single and fixed trigger, which contributes to their robustness. Consequently, when overlaying randomly selected clean images onto backdoored images, the trigger remains effective in the overlaid image, resulting in a strong classification bias. The confidence vector of soft labels tends to favor the target label and exhibits reduced ambiguity. Conversely, for global and sample-specific trigger patterns, the triggers lack the characteristic of being single and fixed. Due to this limitation in the STRIP method's concept, the enhanced method STRIP-Cl can only detect patch-based perturbations and fails to identify global perturbation types.

\subsubsection{Defense of SSL-Cleanse}
SSL-Cleanse\cite{zheng2023ssl} can defend against two types of backdoor attacks targeting SSL encoders: SSL-Backdoor and ESTAS. Both SSL-Backdoor and ESTAS employ patch-based trigger addition, where the attacker modifies a portion of the training data by adding specific triggers to activate the backdoor. SSL-Cleanse\cite{zheng2023ssl} utilizes clustering algorithms and reverse recognition techniques to detect backdoor attacks. By clustering samples generated by the SSL encoder, it can identify groups of samples with anomalous triggers. Then, through reverse pattern recognition techniques, SSL-Cleanse\cite{zheng2023ssl} can identify and remove these triggers, restoring the encoder to a clean state without backdoors.

\begin{table}[htbp]
    \centering
    \caption{The certified accuracy and attack success rates for the backdoored downstream classifiers defended by SSL-Cleanse\cite{zheng2023ssl}. The pre-training dataset is CIFAR10.}
    \resizebox{0.50\columnwidth}{!}{
    \begin{tabular}{cccc}
       \toprule[1.2pt]
          Downstream Dataset & CA & BA & ASR  \\
       \midrule
       STL10  & 76.14  & 24.05 & 7.75  \\
       GTSRB  & 81.84  & 5.90 & 0.00  \\
       SVHN  & 61.72  & 52.11 & 24.13  \\
       \bottomrule[1.2pt]
    \end{tabular}}
    \label{tab: SSL-Cleanse}
\end{table}


\begin{figure*}[htbp]
	\centering
	\begin{minipage}[c]{0.40\textwidth}
		\centering
		\includegraphics[width=\textwidth]{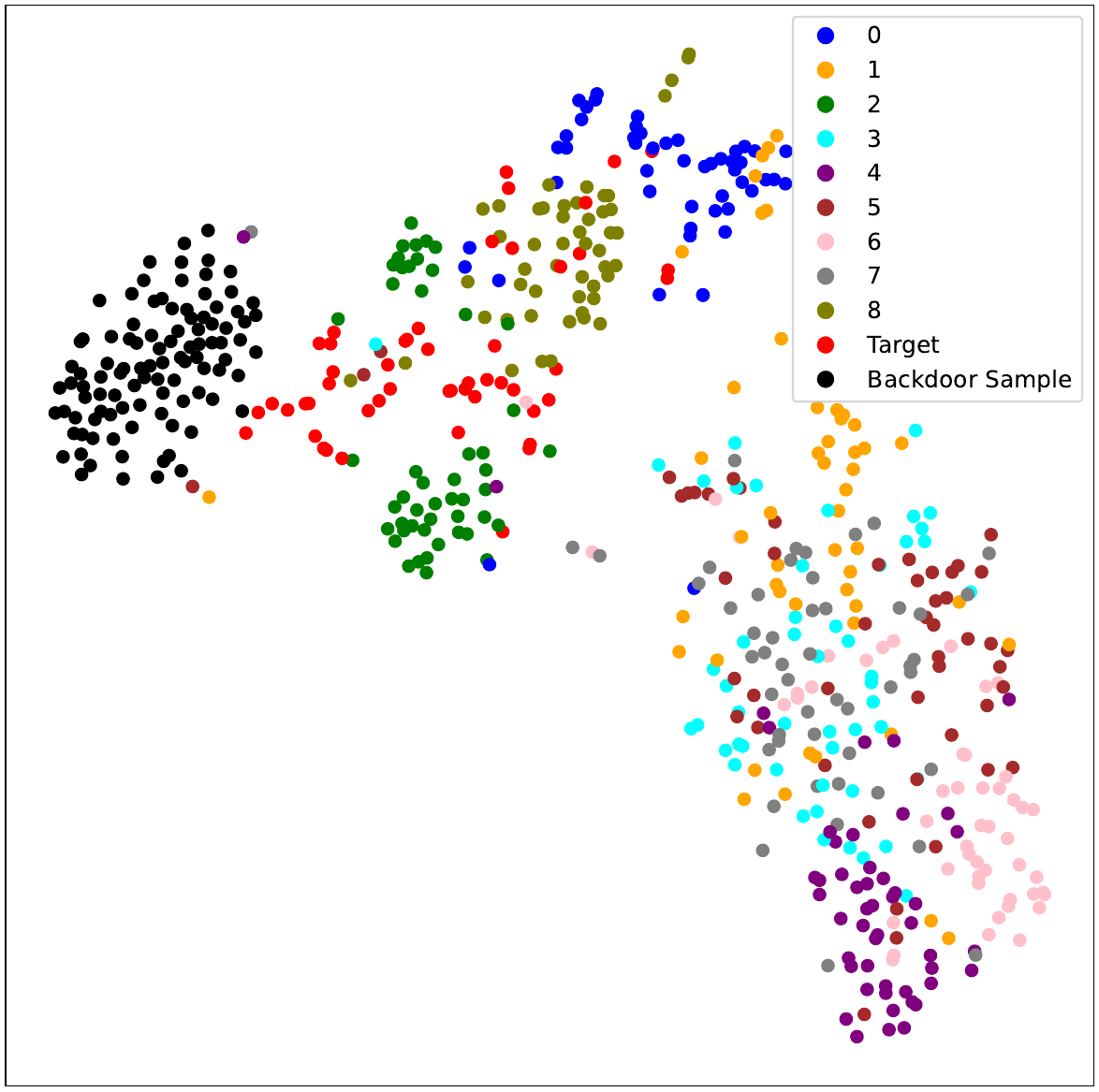}
		\subcaption{}
		\label{fig:8a}
	\end{minipage} 
	\begin{minipage}[c]{0.40\textwidth}
		\centering
		\includegraphics[width=\textwidth]{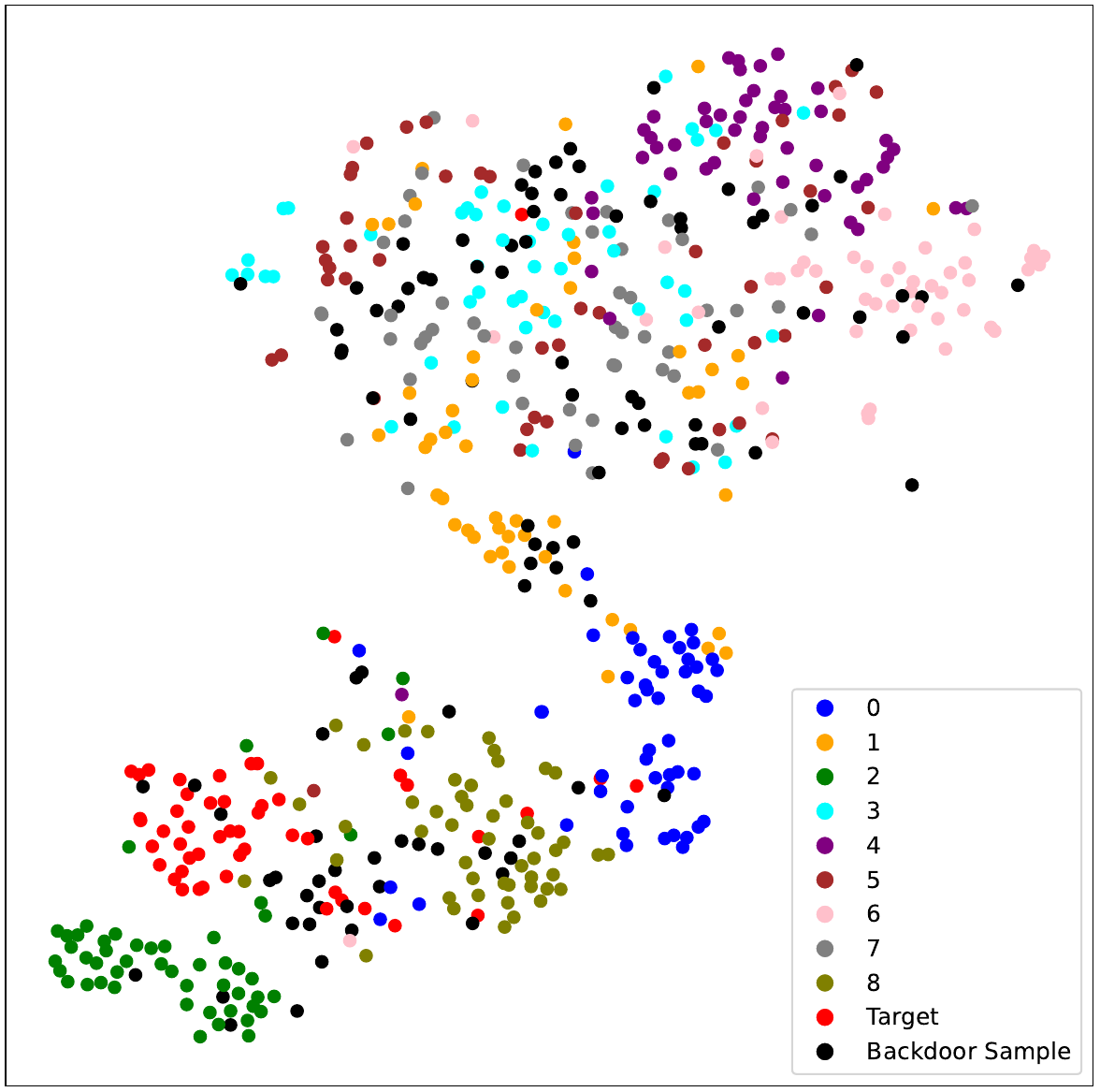}
		\subcaption{}
		\label{fig:8b}
	\end{minipage} \\
	\begin{minipage}[c]{0.40\textwidth}
		\centering
		\includegraphics[width=\textwidth]{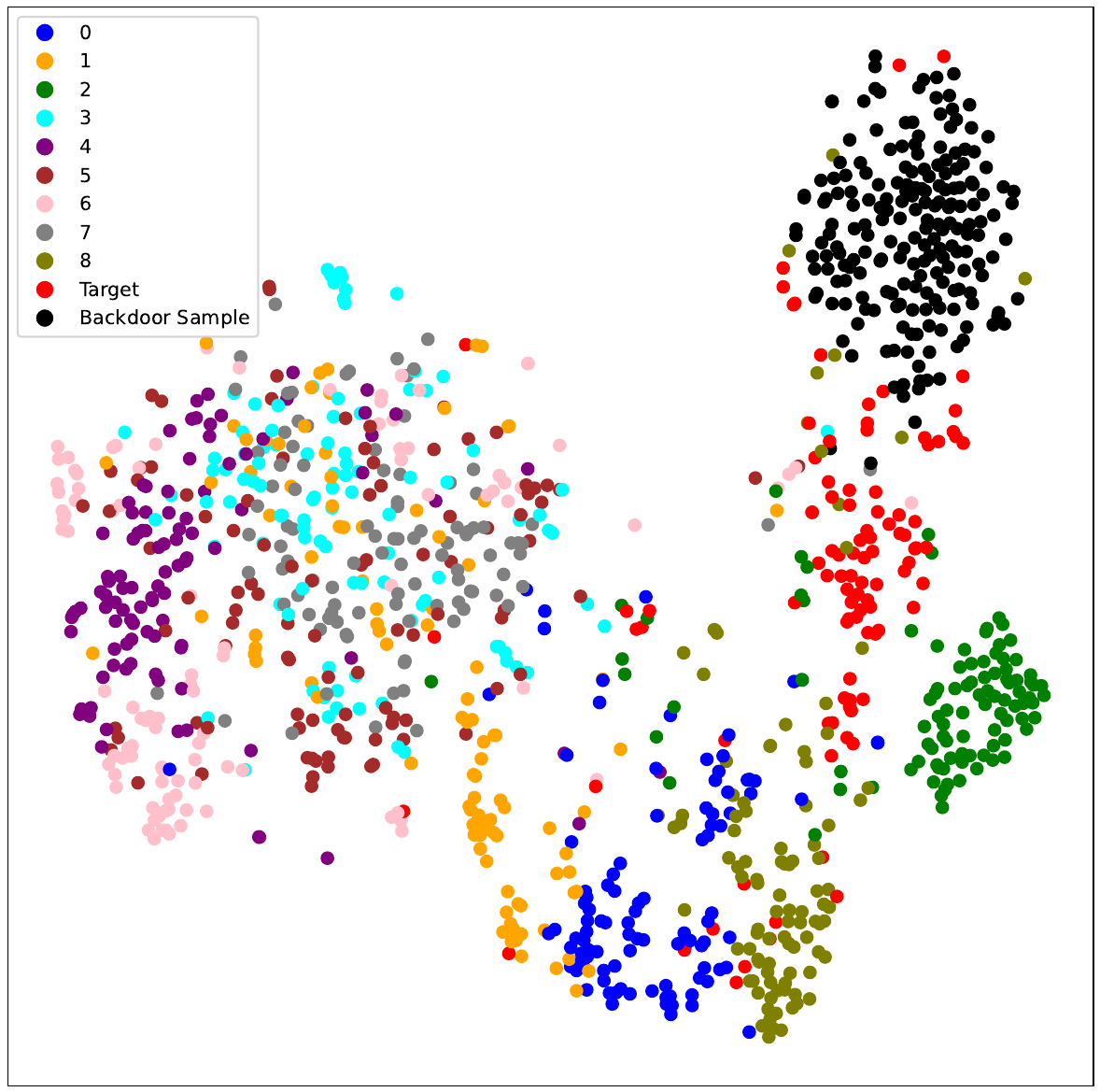}
		\subcaption{}
		\label{fig:8c}
	\end{minipage}
         \begin{minipage}[c]{0.40\textwidth}
		\centering
		\includegraphics[width=\textwidth]{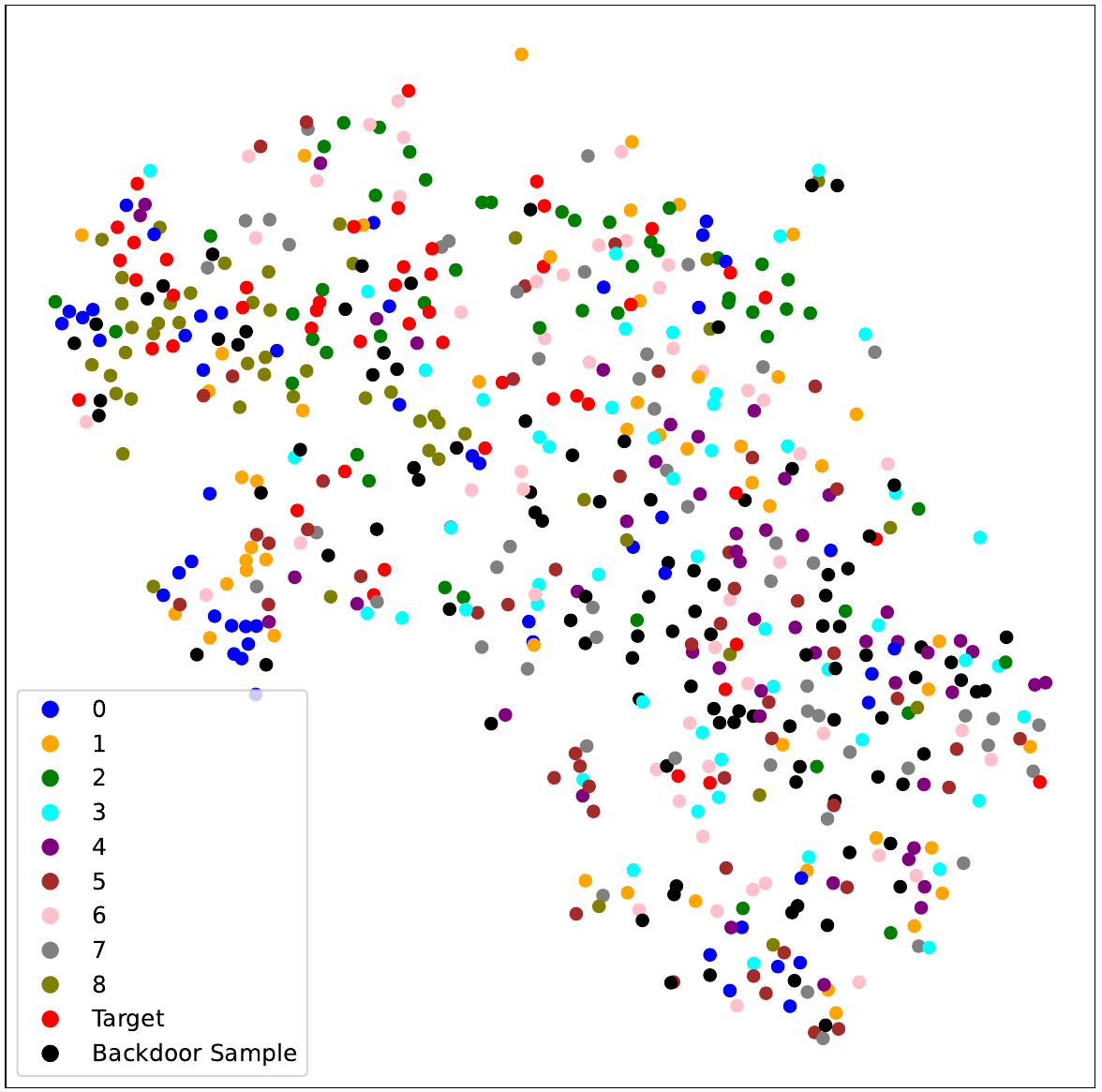}
		\subcaption{}
		\label{fig：8d}
	\end{minipage} 
	\caption{t-SNE visualization of embedded triggers and clean input features in BadEncoder and GhostEncoder attacks (target class input: red; Embedded trigger input: black)
    (a) The results of the BadEncoder before the SSL-Cleanse defense.
    (b) The results of the BadEncoder after the SSL-Cleanse defense.
    (b) The results of the GhostEncoder after the SSL-Cleanse defense.
    (b) The results of the GhostEncoder after the SSL-Cleanse defense.}
	\label{fig:feature-SNE}
\end{figure*}

However, for the GhostEncoder attack proposed in this work, which utilizes global perturbations that affect the entire image space, the trigger has an impact on every pixel, unlike patch triggers that are limited to specific locations. This makes the detection and removal of triggers more challenging, and the sample-specific nature of GhostEncoder makes it difficult to recognize the trigger as a unified pattern, further hindering direct identification and elimination.

In this work, we assume that the triggers used in the reverse-engineered attacks using the proposed defense method are optimal, meaning they are the original triggers from our GhostEncoder technique. These triggers are employed to mitigate the impact of the backdoor in our generated backdoor encoder, resulting in an optimized clean encoder. We conducted two rounds of optimization on the model, and when the upstream dataset was CIFAR10, we defended the backdoor encoders of the downstream datasets STL10, GTSRB, and SVHN, obtaining clean triggers. Table \ref{tab: SSL-Cleanse} presents the test results of the clean triggers on the downstream datasets. It is evident that SSL-cleanse\cite{zheng2023ssl} effectively reduces the ASR of GhostEncoder. However, it significantly compromises the classification performance of the SSL encoder on normal inputs. Therefore, this defense method is not meaningful against GhostEncoder. Consequently, the SSL-Cleanse\cite{zheng2023ssl} method cannot directly address the global perturbation triggers of GhostEncoder attacks.

To corroborate the aforementioned analysis, we conducted tests on the backdoor encoders of BadEncoder and GhostEncoder before and after applying the SSL-Cleanse defense. For this purpose, we employed t-SNE visualization to examine the features of embedded triggers and clean inputs in the test set. Our experimental setup utilized the upstream pre-training dataset CIFAR10 and the downstream dataset STL10, as depicted in Fig. \ref{fig:feature-SNE}. In the case of BadEncoder (Fig. \ref{fig:8a},\ref{fig:8b}), prior to the defense, the clusters of backdoor inputs (in black) and clean inputs from the target class (in red) exhibit a high degree of proximity in the feature space. However, after defense, a significant reduction in the aggregation of backdoor inputs is observed, as they move away from the target class. According to the outcomes presented in Table \ref{tab: SSL-Cleanse}, GhostEncoder, after being subjected to the SSL-Cleanse defense, experiences a substantial decrease in its BA to 24.05\%. This deterioration in classification performance results in a disorganized and unclustered representation as shown in Fig. \ref{fig：8d}.

\subsubsection{Other Defenses}
In addition to the aforementioned defenses, we also explored DECREE \cite{feng2023detecting}, which can detect the presence of backdoors in models. The authors noted that DECREE failed to detect SSL-Backdoor and CTRL\cite{li2022demystifying} attacks with PL1-Norm around 0.23. The reason for the failure to detect SSL-Backdoor was its lower-than-expected ASR ($<10\%$), which falls outside the expected ASR range ($>99\%$). Compared to SSL-Backdoor and CorruptEncoder, our GhostEncoder employs more sample-specific and invisible triggers with higher PL1-Norm. Additionally, GhostEncoder's ASR is not within the expected range of DECREE, making it undetectable by the DECREE defense method.
While DECREE can detect the presence of backdoors, it cannot eliminate them. Therefore, we also explored the PatchSearch defense, which operates during the training phase and assumes poisoning-based backdoor attacks. However, this defense does not align with our threat model in this work since GhostEncoder utilizes weight modification-based attack methods instead.

\section{Discussion}

\subsection{Insights from Experimental Observations}

Under the parameter settings outlined in Section 5.1.2, we trained watermark encoders and observed that the effectiveness of generated watermarks is influenced to some extent by the stochastic nature of the training process. Through extensive experimentation, we elucidated the impact of watermark strength on the performance of attacks on downstream datasets, establishing a connection between this impact and the disparities in data distribution between pretraining and downstream datasets. Specifically, we found that watermark encoders with greater potency are better suited for generating watermark trigger patterns of a certain intensity when the data distribution of the downstream dataset is akin to that of the pretraining dataset. Conversely, in cases where there is significant dissimilarity between the data distributions of upstream and downstream datasets, encoders of a less robust nature are more adept at producing trigger perturbations. Building upon these insights, we conducted experiments wherein we tailored the training of respective watermark encoders suitable for backdoor attacks based on the distinct data distributions of various downstream datasets.

\subsection{Other Settings in self-supervised analysis}

Existing backdoor attacks on SSL are only effective when using patch-based sample recognition triggers. To better understand backdoor attacks in SSL, we apply two existing attacks from supervised learning to our scenario: adversarial sample attack SSAH \cite{luo2022frequency}and poisoning-based ISSBA \cite{2}. Our adversarial sample attack has two settings: one follows the attack procedure of BadEncoder \cite{1} for the Image-on-Image attack used in this work, but replaces the backdoored samples generated by patch-based triggers with adversarial samples; the other is poisoning-based SSAH, where poisoned samples with added global triggers are mixed into the clean training set of the upstream encoder. Poisoning-based ISSBA aligns with the attack setting of poisoning-based SSAH. We then evaluate the ASR of downstream classifiers trained from the backdoored encoder. The results are shown in Table \ref{tab:multiple attacks}.

From the experimental results, it can be observed that adversarial sample attacks and targeted poisoning attacks on pre-trained self-supervised encoders are difficult to succeed. These attacks can be successful and covert in supervised learning because there is a specific target label that provides strong hints during the attack process. However, SSL only considers positive pairs or negative pairs and lacks obvious features (such as patch-based triggers). Therefore, it is challenging for such sample triggers to establish strong correlations between victim images and target images.

\begin{table}[]
    \centering
    \caption{ASR results of multiple attacks.}
    \resizebox{0.7\columnwidth}{!}{\begin{tabular}{ccccc}
       \toprule[1.2pt]
         & SSAH-poison & SSAH-modify & ISSBA-poison & GhostEncoder  \\
       \midrule
       ASR  & 10.15  & 10.79 & 11.62 & 96.57 \\
       \bottomrule[1.2pt]
    \end{tabular}}
    \label{tab:multiple attacks}
\end{table}

\section{Related Work}
\subsection{Backdoor Attack}
The Backdoor attack is a rapidly growing and promising topic in the field of network security, which can have detrimental effects on security by poisoning the training process of DNNs.
In the realm of supervised learning, researchers have extensively explored backdoor attacks targeted at image classifiers. These attacks involve the injection of patches, also known as triggers, into images. By augmenting clean datasets with toxic samples during training, attackers aim to manipulate the model's behavior. Notable works in this domain, including references \cite{gu2017badnets,liu2018trojaning,liu2017neural}, emphasize the model's inclination to activate the backdoor effect exclusively on data containing the embedded triggers. Despite the compromised performance on clean data, models trained with poisoned datasets tend to misclassify images with triggers into the attacker's predefined target class during testing. While earlier studies, like \cite{nguyen2020input}, focus on sample-specific backdoor attacks, they often require meticulous control over training loss and data, diminishing their practical threat level. Moreover, the conspicuousness of trigger patterns, such as the visible 3x3 white square in \cite{gu2017badnets}, renders them susceptible to defense mechanisms employed by state-of-the-art methods.

Concealing the trigger is pivotal for more sophisticated attacks. Chen et al. ~\cite{chen2017targeted} underline the importance of concealing triggers for enhanced stealthiness. They advocate for toxic samples that closely resemble benign ones, thereby making the attack more covert. A covert blending strategy, proposed by Chen et al., directly integrates backdoor triggers into benign images, avoiding blatant patching. Parallel studies explore covert attacks in various contexts. For instance, Quiring et al.~\cite{quiring2020backdooring} address image scaling, while Saha et al.~\cite{saha2020hidden} assume adversaries possess knowledge of the model structure. Authors in \cite{zhang2022poison} exploit image structure to create covert trigger regions and embed them using a deep injection network, achieving concealed attacks. Similarly, in \cite{wang2022invisible}, an RGB to YUV channel transformation combined with DCT conversion is proposed to insert triggers. These methods, while potent, maintain a fixed nature during training and testing.

The domain of SSL introduces distinct perspectives on backdoor attacks. Broadly, these attacks fall into two categories: poisoning-based\cite{carlini2021poisoning,saha2022backdoor,liu2022poisonedencoder,li2022demystifying} and non-poisoning-based attacks\cite{1}. Saha et al.~\cite{saha2022backdoor} introduced poisoning-based backdoor attacks within SSL, wherein adversaries contaminate a limited training dataset. These attacks yield fixed trigger patterns for backdoor samples, resulting in relatively modest success rates. Building on this, Li et al.~\cite{li2022demystifying} proposed a novel poisoning-based attack approach, defining trigger patterns as specific frequency domain perturbations. Despite their potency, these trigger patterns remain static. Notably, backdoor attacks can manifest during various stages, including data collection, training, and deployment. Liu et al.~\cite{1} devised a method for injecting backdoors into pre-trained encoders, altering model weights and associating trigger patterns with target class inputs. Nevertheless, this approach remains reliant on a singular fixed trigger pattern.

\subsection{Backdoor Defense}

Various defense methods have been proposed to counter backdoor attacks. Currently, defense efforts in the context of self-supervised scenarios primarily focus on detecting backdoor models or eliminating maliciously implanted Trojan horse models. To distinguish between backdoor and normal models, existing techniques employ vertical trigger patterns for model decision-making and discernment based on reverse trigger features (such as trigger size)\cite{liu2019abs,shen2021backdoor,tao2022better}. DECREE\cite{feng2023detecting} belongs to the category of defense methods for detecting backdoor models. It is designed for backdoor detection in pre-trained encoders, requiring no classifier head or input labels, but it can only determine whether a model contains a backdoor, without the capability to remove it. Other methods utilize meta-classifiers to detect the presence of backdoors in models\cite{kolouri2020universal,feng2023detecting}; however, these methods entail significant computational resources and rely on a strong assumption—namely, that trigger size is known—thus rendering them less practical.

In the realm of removing tainted samples, Tran et al.~\cite{tejankar2023defending} attempt to cleanse toxic samples by analyzing the spectrum of latent features. Yet, this method assumes complete access to the infected training data, which is impractical in real-world applications. The STRIP method proposed by Gao et al.\cite{strip} involves overlaying various patterns onto suspicious images to filter out malicious samples, and then observing the randomness of their predictions. They assume that the input method of the backdoor trigger is unknown. Additionally, Tejankar et al.~\cite{tejankar2023defending} take a novel approach by focusing on defense during the training phase of encoder models. They mitigate the impact of trigger patterns by searching for trigger-paste positions, effectively eliminating backdoors. Recently, Zheng et al.\cite{zheng2023ssl} introduced a new defense method, SSL-Cleanse, which employs clustering algorithms and reverse identification techniques to detect backdoor attacks. By clustering samples generated by the SSL encoder, it identifies clusters of anomalous triggers and subsequently employs reverse pattern recognition techniques to identify and eliminate these triggers. This work also delves into the impact of this defense method on GhostEncoder.

\subsection{Self-supervised Learning}
Modern computer vision systems excel in various challenging computer vision tasks, largely relying on the availability of large annotated datasets. However, acquiring such data is both time-consuming and expensive, making it impractical to construct large-scale labeled datasets in many real-world scenarios. Therefore, devising learning methods that can successfully recognize new concepts with only a small number of labeled examples is a crucial research challenge.

SSL is a general framework that aims to train image encoders from large amounts of unlabeled data, extracting useful information directly from the data itself, unlike supervised learning which requires manually labeled data. Among many methods, such as MoCo \cite{8}, SimCLR\cite{9}, Sim-CLRv2\cite{10}, and CLIP\cite{11}, contrastive learning achieves state-of-the-art performance. This method compares samples with semantically similar examples (positive pairs) and those with semantically dissimilar ones (negative pairs). By designing the model structure and comparison loss, the positive pairs with similar semantics are represented closer in the representation space, while the semantically dissimilar negative pairs correspond to farther representations, achieving a similar clustering effect.

In the field of computer vision, SSL has numerous applications. Doersch et al.\cite{3} proposed a method for training a convolutional neural network (CNN) by predicting the relative position of two randomly sampled, non-overlapping image patches. Later, papers in \cite{4,5} extended this idea to predicting the permutation relationship between multiple randomly sampled and permuted blocks. Besides patch-based methods, some techniques utilize image-level self-supervised losses. For instance, in \cite{6}, authors proposed using the colorization of grayscale images as an auxiliary task. Another example of a pretext task is in \cite{7}, which predicts the angle of the rotation transformation applied to the input image.

\section{Conclusion}

In this work, we propose the first dynamic covert backdoor attack against SSL, known as GhostEncoder. We demonstrate that existing backdoor attacks are susceptible to mitigation by current defense mechanisms primarily due to the visibility or static nature of their trigger mechanisms. Building on this understanding, we explore a novel trigger pattern and apply it for the first time to non-poisoned backdoor attacks.Experimental results confirm that our GhostEncoder deceives victim models with a high attack success rate while maintaining model utility. Furthermore, it is evident that the existing defenses against backdoor attacks are insufficient to counter our approach. We investigate the reasons behind STRIP's failure in the self-supervised setting and propose a novel enhancement, termed STRIP-Cl.Interesting avenues for future research encompass: 1) Extending our attack to other domains within self-supervised scenarios, such as natural language processing; 2) Devising novel defense strategies to thwart our attack.

\section*{Acknowledgments}
We thank anonymous reviewers for their constructive
comments to this work. The work has been supported by the National Key R\&D Program of China (Grant No.2021YFB3100700), the National Natural Science Foundation of China (No. U22B2029, 62272228) and Shenzhen Science and Technology Program (Grant No.JCYJ20210324134408023).

\def\shortbib{0}
{\small
\bibliographystyle{elsarticle-num-names}
\bibliography{myref}
}
\end{document}

%% file: header/header-Fixed.tex
\numberwithin{equation}{section}

\renewcommand{\epsilon}{\varepsilon}

%% file: header/header-diy.tex
\newcommand{\msg}{m\xspace}